\documentclass[10pt]{article} 
\usepackage[preprint]{tmlr}

\usepackage[most]{tcolorbox}
\usepackage{xcolor}

\usepackage{url}

\usepackage{wrapfig}
\usepackage{amsmath}
\usepackage{amssymb}
\usepackage{amsthm}
\usepackage{algorithm}
\usepackage{algpseudocode}

\DeclareMathOperator*{\argmin}{arg\,min}
\usepackage{hyperref}
\usepackage[capitalise]{cleveref}
\usepackage{refcount}

\newcommand{\X}{\mathcal X}
\newcommand{\Y}{\mathcal Y}

\newcommand{\G}{\mathcal G}
\newcommand{\F}{\mathcal F}
\newcommand{\Lo}{\mathcal L}
\newcommand{\D}{\mathcal D}
\newcommand{\R}{\mathbb R}
\newcommand{\E}{\mathbb E}
\newcommand{\U}{\mathcal U}
\newcommand{\dm}{^{\mathrm{DM}}}

\newcommand{\Feg}{\mathcal{F}^{\mathrm{eg}}}
\newcommand{\Frw}{\mathcal{F}^{\mathrm{rw}}}
\newcommand{\pre}{_{\textrm{pre}}}
\newcommand{\bm}{b^-}
\newcommand{\bp}{b^+}

\newcommand{\thelam}[1][\lambda]{\theta_{(#1)}}
\newcommand{\lpre}[1][\theta]{\mathcal{L}_{\mathrm{pre}}(#1)}
\newcommand{\lpost}[1][\theta]{\mathcal{L}_{\mathrm{post}}(#1)}
\newcommand{\fpre}[1][\theta]{\mathcal{F}_{\mathrm{pre}}(#1)}
\newcommand{\fpost}[1][\theta]{\mathcal{F}_{\mathrm{post}}(#1)}
\newcommand{\frwpre}[1][\theta]{\mathcal{F}_{\mathrm{pre}}^{\mathrm{rw}}(#1)}
\newcommand{\frwpost}[1][\theta]{\mathcal{F}_{\mathrm{post}}^{\mathrm{rw}}(#1)}
\newcommand{\fegpre}[1][\theta]{\mathcal{F}_{\mathrm{pre}}^{\mathrm{eg}}(#1)}
\newcommand{\fegpost}[1][\theta]{\mathcal{F}_{\mathrm{post}}^{\mathrm{eg}}(#1)}
\newcommand{\upre}[2][\theta]{\mathcal{U}^{#2}_{\mathrm{pre}}(#1)}
\newcommand{\upost}[2][\theta]{\mathcal{U}^{#2}_{\mathrm{post}}(#1)}
\newcommand{\lpreg}[2][\theta]{\mathcal{L}^{#2}_{\mathrm{pre}}(#1)}
\newcommand{\lpostg}[2][\theta]{\mathcal{L}^{#2}_{\mathrm{post}}(#1)}

\newcommand{\C}{C}
\newcommand{\Cacc}{C^{\mathrm{acc}}}
\newcommand{\Csel}{C^{\mathrm{sel}}}
\newcommand{\B}{B}
\newcommand{\Brej}{B^{\mathrm{rej}}}
\newcommand{\Bfn}{B^{\mathrm{fn}}}
\newcommand{\dsc}{\zeta}

\theoremstyle{plain}
\newtheorem{theorem}{Theorem}[section]
\newtheorem{proposition}[theorem]{Proposition}
\newtheorem{lemma}[theorem]{Lemma}

\theoremstyle{definition}
\newtheorem{definition}[theorem]{Definition}
\newtheorem{assumption}[theorem]{Assumption}
\theoremstyle{remark}

\crefname{assumption}{Assumption}{Assumptions}
\crefname{claim}{Claim}{Claims}

\definecolor{mypink}{HTML}{de77ae}
\definecolor{group0Color}{HTML}{8da0cb}
\definecolor{group1Color}{HTML}{fc8d62}
\definecolor{dominantColor}{HTML}{7570b3}
\definecolor{dominatedColor}{HTML}{e7298a}
\definecolor{lightgrey}{RGB}{240, 240, 240}
\definecolor{darkgrey}{RGB}{120, 120, 120}
\definecolor{myBrown}{HTML}{a6761d}

\title{The Delayed Societal Cost of Unfairness}

\author{\name Giovani de Almeida Valdrighi \email giovani.valdrighi@ic.unicamp.br \\
      \addr Instituto de Computação \\
  Universidade Estadual de Campinas \\
  Campinas, Brazil \\
      \AND
      \name Isabel Valera$^{\dagger}$  \email ivalera@cs.uni-saarland.de\\
      \addr Department of Computer Science \\
  Saarland University \\
  Saarbrücken, Germany \\
      \AND
      \name Marcos Medeiros Raimundo$^{\dagger}$ \email mraimundo@ic.unicamp.br \\
      \addr  Instituto de Computação \\
  Universidade Estadual de Campinas \\
  Campinas, Brazil \\}

\def\month{MM}  
\def\year{YYYY} 
\def\openreview{\url{https://openreview.net/forum?id=XXXX}} 

\begin{document}

\maketitle

{\renewcommand{\thefootnote}{}%
 \footnotetext{$^{\dagger}$Equal advising.}}

\begin{abstract}
Fair machine learning typically evaluates the \textit{societal costs} of a decision policy—the loss incurred by the decision-maker (DM) and the unfair distribution of costs among demographic groups of decision-subjects (DS)—\textit{on static data}, where reducing one often comes at the expense of the other.
Yet decisions change the populations on which they act, and prior work has shown that fairness interventions designed on static data may fail to benefit, or even harm, the disadvantaged group over time.
We investigate a complementary question: \textit{When can a policy that is Pareto-optimal on pre-deployment data be outperformed by a fairer policy once each is evaluated on the population it induces, reducing costs for one stakeholder without increasing them for the other?}
In a setting where qualification scores evolve in response to binary decisions and their outcomes, we show that under Rawlsian fairness, which never lowers any group's acceptance rate, placing sufficiently high weight on fairness reduces societal costs whenever accepting disadvantaged DS at the DM's break-even point leaves them in a state less costly to the DM than rejecting them. 
Under egalitarian fairness, which may lower the advantaged group's acceptance rate, this additionally requires the reduction in DM loss on the disadvantaged group to outweigh the increase on the advantaged group.
Under both fairness notions, however, when DM loss is accuracy-based, improvements in the post-deployment population may arise from declining DS qualification, revealing that gains for the DM do not necessarily translate into better outcomes for DS. 
Our results show that \textit{policy-induced population dynamics can overturn apparent performance–fairness trade-offs, allowing fairer policies to reduce the delayed societal costs of decision-making.}

\end{abstract}

\section{Introduction}

Algorithmic decision-making in consequential applications such as lending, hiring, and school admission imposes \emph{societal costs}: decisions cost not only the decision-maker (DM) but also the decision-subjects (DS), and these costs can be unevenly distributed across demographic groups. This motivates a multi-objective perspective in which, alongside DM loss, we introduce the (un)fairness of decisions as a secondary objective, reflecting DS interests~\citep{gupta2026first}. Since no single policy usually minimizes both objectives, we draw our attention to \emph{Pareto-optimal} ones, where one objective cannot be improved without worsening the other. Such policies reveal the attainable trade-offs, i.e., how much DM loss increases to reach a given fairness level~\citep{wei2022fairness,menon2018cost}, and allow stakeholders to discuss which policy to deploy. However, optimality is only defined on the static data on which policies are evaluated.

Decisions reshape the population they act on; e.g., a hired employee gains experience that strengthens future application. Prior work has shown that, because of these effects, policies that are fair on static data may fail to remain fair, or may even harm DS, over time~\citep{liu2018delayed,damour2020fairness}. Methods that account for decision effects require how it influences DS~\citep{rateike2024designing,wen2021algorithms}, which can be highly complex, or learning by exploration~\citep{yin2023long, kilbertus2020fair}, which can raise ethical concerns in certain applications, e.g., a bail decision based on a coin flip~\citep{kilbertus2020fair}. Thus, commonly, researchers still assess trade-offs between performance and fairness on a historical, static dataset, even though each policy later faces societal costs from the population it shapes. 

\begin{figure}[t]
    \centering
    \includegraphics[width=0.75\linewidth]{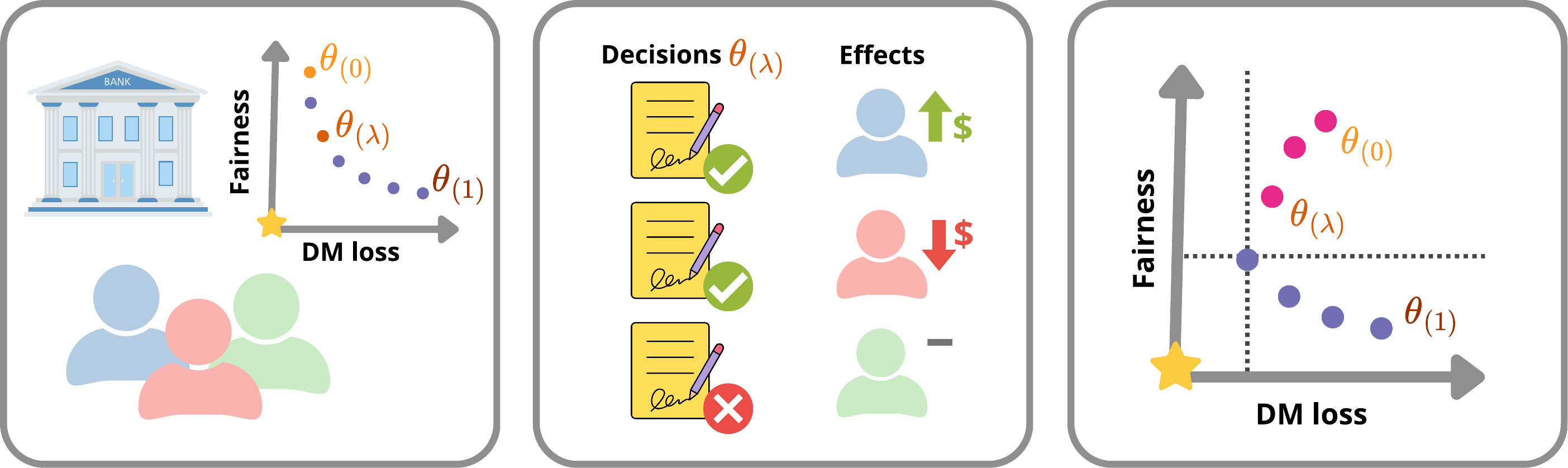}
    \caption{Illustration of the decision-making process in which stakeholders (DM and DS) assess the optimal trade-offs between performance and fairness, agree on a policy $\thelam$, and deploy it. Once deployed, the policy supports decisions that will reshape the population it acts on. Over time, policies incur costs from the updated distribution and may fail to remain optimal compared to other pre-deployment Pareto-optimal solutions.    
    }
    \label{fig:movation}
\end{figure}

Throughout this paper, we use lending as a motivating example (\cref{fig:movation}). We consider DS from two demographic groups that apply for a loan, and a DM aims to accept applicants with a high chance of repayment. Because qualification can be unevenly distributed across groups, optimizing only the DM loss yields policies that serve one group poorly. The DM collects policies $\thelam$ with different trade-offs by optimizing $(1-\lambda)\,\texttt{performance} + \lambda\, \texttt{fairness}$ for different fairness weights $\lambda$, and stakeholders agree on one policy from these options. Once deployed, decisions affect DS; for example, a loan can enable professional training and a raise, or become burdensome debt. When applicants return, the same policy faces this updated population. Hence, its DM loss and fairness depend not only on its decisions but also on how it shapes the population it interacts with.

In this context, we ask whether the trade-off agreed on before deployment on a static dataset persists after it: \emph{can~a~fairer~policy $\thelam[\mu]$, with $\mu > \lambda$, dominate $\thelam$ after deployment, i.e., be at least as good on both objectives and strictly better on one when each policy is evaluated on the population it induces?} We formalize this question within performative prediction~\citep{perdomo2020performative} and restrict the comparison to Pareto-optimal policies on the static data, since these are the ones stakeholders choose before deployment. Dominance means that a higher fairness weight would have been better for both stakeholders. We first show that dominance occurs when the gains from the fairer policy from reshaping the population offset its static suboptimality in DM loss while maintaining its fairness advantage. We then specialize this condition to lending dynamics adapted from \citet{liu2018delayed}, covering accuracy and selective DM costs, DS costs that recover demographic parity~\citep{dwork2012fairness} and equality of opportunity~\citep{hardt2016equality}, and two different theories of justice, Rawlsian~\citep{rawls1971theory} and egalitarian~\citep{arneson1999egalitarianism}. The summary of our findings is:
\begin{enumerate}
    \item Under Rawlsian fairness, a higher fairness weight never lowers a group's acceptance rate nor worsens post-deployment fairness, so dominance depends only on the DM loss. If accepting disadvantaged DS at the DM's break-even point leaves them in a state cheaper for the DM than rejecting them, every weight below some $\lambda^\dagger > 0$ is dominated (\cref{theo:main}).
    \item Under egalitarian fairness, which can lower the advantaged group's acceptance rate, dominance additionally requires the DM gain on the disadvantaged group to outweigh its loss on the advantaged group.
    \item When DM loss is based on the accuracy metric (i.e., defined by false positive/negative costs), post-deployment gains can come from lowering DS qualification, so a DM gain does not imply DS are better off.
    \item On FICO data, small fairness weights are dominated under selective costs, and the minimal weight that produces a dominant solution grows as repayments raise scores more relative to how much defaults lower them.
\end{enumerate}

\section{Related Work}

\subsection{Performance and Fairness Trade-offs}

In this work, we focus on \textit{group fairness}, one of the most common fairness principles in machine learning, which measures unfairness by how unevenly costs are distributed across demographic groups~\citep{mehrabi2021survey}. In this setting, how ``unfairness’’ is measured can also differ; e.g., egalitarian fairness~\citep{arneson1999egalitarianism} aims to equalize costs across groups while Rawlsian fairness~\citep{rawls1971theory} instead minimizes the loss of the worst-off group. A large body of work has been produced in learning fair predictors, using, for example, constrained optimization, representation learning, and post-processing~\citep{mehrabi2021survey, zafar2017fairness, zemel2013learning, hardt2016equality}. A key distinction lies between constrained and multi-objective perspectives. While in constrained fairness (e.g., \citep{zafar2017fairness, cruz2023fairgbm}), the goal is to obtain a classifier that minimizes DM loss with an upper limit on unfairness, multi-objective instead describes the set of attainable trade-offs, quantifying how much DM loss increases for each fairness level~\citep{menon2018cost, wei2022fairness, gupta2026first}. 

Prior work has debated whether this trade-off is inevitable. Some works argue that this occurs when qualifications differ across groups~\citep{menon2018cost, zhao2022inherent}, while others argue that this results from biased labels against the disadvantaged group~\citep{wick2019unlocking, dutta2020there}. In this work, we argue that conflict between DM loss and fairness can also stem from evaluating policies under a static distribution, while such trade-offs might not persist once deployed.

\subsection{Fairness in Dynamic Settings}

Research on \textit{long-term fairness} studies fair decision-making when decisions reshape the population they act on; for a complete survey, see \citet{gohar2025longterm}. A central finding is that imposing fairness at each round may fail to ensure fairness over the full period, or even harm the disadvantaged group~\citep{liu2018delayed,damour2020fairness,zhang2020fair,mishler2022fair,zhang2019group}. When DM knows the dynamics between decisions and DS distribution, or when random exploration is possible during learning, prior work has proposed reinforcement learning algorithms to learn long-term fair policies~\citep{yin2023long, yu2022policy, rateike2024designing, wen2021algorithms, raab2024fair}. One interesting perspective is how such dynamics interact with common fair results~\citep{somerstep2024algorithmic, jung2020fair}. For example, \citet{somerstep2024algorithmic} showed that the decision effects can break the incompatibility among common fairness metrics~\citep{chouldechova2017fair,kleinberg2017inherent}. 

In this work, we use the \textit{Performative Prediction}~\citep{perdomo2020performative} modeling approach, in which each parametrized model induces a specific distribution observed only after deployment. Prior work has already connected performative prediction to long-term fairness. \citet{hu2022achieving} used it to learn a policy that controls for both short- and long-term fairness, while \citet{jin2026addressing} proposed smooth losses that allow repeated retraining to reduce unfairness. To the best of our knowledge, however, no work has addressed the relationship between dynamics and the performance-fairness trade-off.

\section{Background: Performance and Fairness Multi-Objective Optimization}
\label{sec:background}

Following \citet{gupta2026first}, we frame decision-making as a multi-objective optimization problem in which the decision-maker (DM) balances its institutional loss against the collective burdens experienced by decision-subjects (DS).

\paragraph{Modeling and Notation} DS are represented by variables $(X, Y, G)$, where $X \in \X \subseteq \R$ is a scalar score (e.g., an uncalibrated credit score), $Y \in \Y := \{0, 1\}$ is an outcome label (e.g., loan payment), and $G \in \mathcal G:= \{0, 1\}$ indicates membership in a salient demographic group (e.g., defined by gender). We denote by $\rho^g:= P(G = g) > 0$ the proportion of group $g$, by $\D$ the marginal distribution of $X$, and the probability of a positive label given score $x$ and group $g$ by  $\alpha(x, g):= P(Y = 1 \mid X =x, G =g)$, which we call qualification. We also write $\mathbf{1}\{\cdot\}$ for the indicator function. Although simplified, this formulation permits us to isolate the trade-off between performance and fairness~\citep{liu2018delayed, gupta2026first, wilms2026fairness}.

\subsection{Decisions and Institutional Loss}

The DM issues a binary decision $D \in \{0, 1\}$ for each DS, where $D= 1$ denotes acceptance. Decisions follow a group-specific threshold\footnote{Group-specific thresholds allow the DM to account for baseline disparities in the distribution of scores across groups \citep{hardt2016equality}.} policy $\pi_\theta(x, g):= 1 \{ x \ge \theta^g\}$ where $\theta^g \in \X$. As thresholds fully determine decisions, we will refer to $\theta$ as the policy. The DM aims to minimize the loss $\Lo\dm$: 
\begin{align}
\label{eq:dm_loss}
\Lo\dm(\theta) = \E\left [\sum_{y \in \{0, 1\}} \sum_{d \in \{0, 1\}}\C_{dy} \mathbf{1} \{ Y = y\} 1 \{D = d \} \right ]
\end{align}
Where $\C_{dy}$ is the cost for decision $d$ and label $y$. We assume that $\C_{11} \le \C_{10}$ and $\C_{00} \le \C_{01}$; i.e., correct decisions are never costlier than incorrect ones. In this work, we focus on two definitions of the cost matrix:
\begin{align}
\Cacc = \begin{pmatrix} 0 & c_{FN} \\ c_{FP} & 0 \end{pmatrix}
\qquad
\Csel = \begin{pmatrix} 0 & 0 \\ c_{FP} & c_{TP} \end{pmatrix}
\end{align}
with $c_{FN}, c_{FP} > 0$ and $c_{TP} < c_{FP}$. The \emph{accuracy cost} $\Cacc$ penalizes both false negative and false positive errors. In contrast, the \emph{selective cost} $\Csel$ has no cost for rejected DS and suits settings where the label of rejected DS is never observed \citep{lakkaraju2017selective}, such as lending or hiring. Both matrices accommodate asymmetric costs, which arise naturally in these applications, e.g., in lending, a default ($c_{FP}$) costs more than a repaid loan earns ($c_{TP}$) \citep{moscato2021benchmark}.

\citet{corbett2017algorithmic} showed that threshold policies achieve the optimal DM loss, with thresholds determined by the conditional distribution $\alpha$ and the cost matrix $\C$, independently of the marginal distribution of scores. This policy accepts a DS described by $(x, g)$ whenever the expected cost of acceptance $\C_1(x, g)$ (averaging over $Y$) is lower than the expected cost of rejection $\C_0(x,g)$. Both costs depend on the probability of the positive label $\alpha(x, g)$:  
\begin{align}
\label{eq:decision_costs}
\C_1 (x, g) = \alpha(x, g)(\C_{11} - \C_{10}) + \C_{10} \qquad \C_0 (x, g) = \alpha(x, g)(\C_{01} - \C_{00}) + \C_{00} 
\end{align}
The optimal policy reduces to a threshold policy when the score is informative, meaning a higher $x$ implies a higher probability of $y = 1$, and the optimal threshold is unique when every score in $\X$ can occur. Formally: 

\begin{assumption}
\label{as:outcome_model} 
For each $g \in \G$, $\alpha(\cdot, g)$ is continuous and strictly increasing on $\X$, and the density of $X \mid G = g$ is continuous, bounded, and strictly positive on $\X$.
\end{assumption}

\begin{proposition}[Adapted from \citet{corbett2017algorithmic}]
\label{prop:optimal_dm}
Let $\eta := (\C_{01} - \C_{11}) + (\C_{10} - \C_{00})$ and assume $\eta > 0$. Define $\overline c := (\C_{10} - \C_{00}) / \eta \in [0, 1]$ and assume $\overline c$ lies in the range of $\alpha(\cdot, g)$ for each $g$. Under \cref{as:outcome_model}, the unique minimizer of the DM loss is the threshold policy with
\begin{align}
\theta^g := \inf\{x \in \X : \alpha(x, g) \ge \overline c\}, \quad \text{i.e.,} \quad \alpha(\theta^g, g) = \overline c.
\end{align}
\end{proposition}

\begin{proof}[Sketch Proof]
The DM loss decomposes into the loss evaluated on each group times the group proportion. Since each term depends only on the group-specific threshold $\theta^g$, we can optimize each one separately. For a DS described by $(x, g)$, the DM should accept if and only if $\C_1(x, g) \le \C_0(x,g)$. From \cref{eq:decision_costs}, this inequality gives $\alpha(x, g) \ge \overline c$.
\end{proof}

In simple terms, this result shows that a DM optimizes its own loss by accepting DS only if they have qualification above $\overline c$, i.e., $\alpha(x, g) \ge \overline c$. In the following sections, we call $\overline c$ the break-even point.

\subsection{The Collective Burden of Decisions and Fairness}

Decisions also impose costs on DS, and we measure fairness as how these burdens are distributed across demographic groups. We therefore introduce the DS cost matrix, where $\B_{dy}$ is the cost to a DS of decision $d$ and label $y$, and define the loss of group $g \in \G$ as
\begin{align}
\U^g(\theta) = \E \left [\sum_{y \in \{0, 1\}} \sum_{d \in \{0, 1\}}\B_{dy} \mathbf{1} \{ Y = y\} 1 \{D = d \} | G = g\right ]
\end{align}
A fairness measure $\F : \X \times \X \to \R$ then summarizes the burden of the two groups through $\U^1(\theta)$ and $\U^0(\theta)$, with lower values indicating fairer decisions. This framework recovers common group fairness notions by choosing the DS cost $\B$ and fairness measure $\F$. Viewing acceptance as access to a resource, so that rejection is the costly outcome for DS, we focus our study on two cost matrices:
\begin{align}
    \Brej = \begin{pmatrix} \dsc & \dsc \\ 0 & 0 \end{pmatrix}
\qquad
\Bfn = \begin{pmatrix} 0 & \dsc \\ 0 & 0 \end{pmatrix}
\end{align}
where $\dsc > 0$. The \emph{rejection cost} $\Brej$ penalizes every rejection\footnote{We keep the same cost for all rejections, which recovers demographic parity. Our analysis extends to cases where false negatives and true negatives have different costs, as presented in \cref{app:general_rejection}.}, whereas the \emph{false negative cost} $\Bfn$ penalizes a rejection only when the DS would have had a positive label. We study two definitions of $\F$: (1) \textit{egalitarian} fairness, $\Feg(\theta) := | \U^1(\theta) -  \U^0(\theta)|$, which seeks to equalize the DS group losses, and (2) \textit{Rawlsian} fairness, $\Frw(\theta) := \max \{\U^1(\theta), \U^0(\theta)\}$, which seeks to minimize the loss of the worst-off group. Combined with $\Feg$, the rejection cost recovers demographic parity~\citep{dwork2012fairness}, and the false negative cost yields a variant of equality of opportunity~\citep{hardt2016equality}.

\subsection{Multi-Objective Optimization}
The DM must address the conflict between DM loss and fairness, as policies that optimize each objective will commonly differ. This can be expressed as a multi-objective optimization problem:
\begin{align}
    \min_{\theta \in \Theta}   \left ( \Lo\dm(\theta), \; \;  \F(\theta) \right )
\end{align}
We are interested in Pareto-optimal policies, where no objective can be improved without worsening the other. Formally, a policy $\theta$ dominates another policy $\phi$, written as $\theta \prec \phi$, when $\Lo\dm(\theta) \le \Lo\dm( \phi) $ and $\F( \theta) \le \F( \phi)$ and at least one of the inequalities are strict. Then $\theta$ is Pareto-optimal if no other policy $\phi$ dominates it.

We can obtain a set of Pareto-optimal policies by transforming the multi-objective problem into a single objective parameterized by a \textbf{fairness weight} $\lambda \in [0, 1]$, which sets how much priority the policy gives to fairness relative to the DM loss. The scalarized objective\footnote{There are different approaches to construct the scalarized objective (e.g., the non-linear  Tchebycheff scalarization~\citep{bowman1976relationship}). For simplicity, we focus our analysis on linear scalarizations.} and the respective optimal policy are:
\begin{align}
\label{eq:scalarization}
    \Lo_\lambda( \theta) := (1-\lambda)\Lo\dm( \theta) + \lambda \F( \theta) \qquad  \thelam \in \argmin_{\theta \in \Theta} \Lo_\lambda(\theta)
\end{align}
We define the Pareto-optimal (PO) set $\mathcal{S}:= \{\thelam \mid \lambda \in [0, 1]\}$, as for all $\lambda \in [0, 1]$, the minimizer $\thelam$ is a Pareto-optimal solution\footnote{To ensure that the edge cases $\lambda \in \{0, 1\}$ are also Pareto-optimal, we define $\thelam$ among the policies that minimize the objective as the one that minimizes the other.}. Together with their respective DM loss and fairness, these policies reveal to stakeholders (DM and DS) the attainable trade-offs between the objectives, i.e., how much DM loss must be sacrificed to achieve a given level of fairness, and vice versa. The policies $\thelam[0]$ and $\thelam[1]$ represent edge cases where only the DM loss or the fairness is considered, respectively, and $\thelam[0]$ has the value given in \cref{prop:optimal_dm}.

\section{Problem Formulation: Current Optimality May Lead to Suboptimal Futures}

\begin{wrapfigure}{r}{0.5\textwidth}
  \begin{center}
    \includegraphics[width=0.48\textwidth]{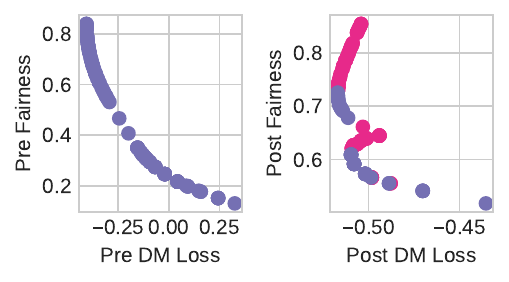}
  \end{center}
  \caption{DM loss and fairness of pre-deployment PO policies $\thelam$ evaluated on pre- and post-deployment distributions. Points in {\color{dominantColor} purple} are {\color{dominantColor} non-dominated}, {\color{dominatedColor} pink} displays {\color{dominatedColor} dominated} policies.}
  \label{fig:pf_pre_post}
\end{wrapfigure}
Decisions alter the population they act on; for example, an approved loan can finance professional training and raise a credit score. Over time, the same policy $\theta$ interacts again with a DS shaped by past decisions. For this reason, a policy optimized in a pre-deployment distribution $\D\pre$ will yield different DM loss and fairness when deployed, as decisions move DS to an induced distribution $\D(\theta)$. 

This framework has been formalized as \textit{Performative Prediction}~\citep{perdomo2020performative}, which we adapt to the multi-objective formulation in \cref{fig:movation}. DM and DS assess the trade-offs among performance and fairness obtained from the
\textbf{pre-deployment PO set} $\mathcal S$, where each policy $\thelam$ is the minimizer in the pre-deployment distribution $\D\pre$ given a fairness weight $\lambda$. The stakeholders agree on a policy $\thelam$, which is then deployed to support decisions and influence the population's scores. We then evaluate this same policy on the induced post-deployment distribution $\D(\thelam)$. However, another policy $\theta_{(\mu)}$ from the pre-deployment PO set $\mathcal{S}$ can dominate $\theta_{(\lambda)}$ when objectives are evaluated on the distribution it induces $\D(\thelam[\mu])$. \cref{fig:pf_pre_post} exemplifies this, where a subset of policies is dominated under the post-deployment distribution.

\begin{definition}[Pre-deployment and post-deployment losses]
Define as $\lpre$ and $\lpost$ the DM loss evaluated on the pre-deployment distribution $\D\pre$ and on the induced post-deployment distribution $\D(\theta)$, respectively. Similarly, define $\upre{g}$ and $\upost{g}$ as the DS loss for group $g$ under the respective distributions, and use them to compute fairness $\fpre$ and $\fpost$.
\end{definition}

\begin{definition}[Post-deployment dominance within pre-deployment PO set $\mathcal S$]
\label{def:dominance}
A fairness weight $\mu \in [0, 1]$ post-deployment dominates another weight $\lambda$ if $\lpost[{\thelam[\mu]}] \le \lpost[{\thelam}]$ and $\fpost[{\thelam[\mu]}] \le \fpost[{\thelam[\lambda]}] $ with at least one strict inequality.
\end{definition}

A dominated fairness weight is a poor choice for both stakeholders: although $\thelam$ is Pareto-optimal on $\D\pre$, deploying $\thelam[\mu]$ instead would lower the DM loss, improving fairness, or both, at no cost to either. This is possible because each policy is evaluated on the distribution it induces, so the trade-off observed before deployment need not persist after it. In particular, a fairer policy may also be better for the DM. The following proposition (proof in \cref{app:proof_dominance_general}) characterizes exactly when this occurs.

\begin{proposition}
\label{prop:dominance_general}
Let $\delta\dm(\theta):= \lpost - \lpre$ be the loss deterioration due to the distribution effect of the policy $\theta$ on the DM loss and $Q\dm (\theta, \phi):= \lpre - \lpre[\phi]$ be the additional cost of policy $\theta$ in the pre-deployment distribution in comparison to another policy $\phi$. $\delta^\F$ and $Q^\F$ are defined analogously for fairness. With two fairness weights $\lambda < \mu$, we have that $Q\dm(\thelam[\mu],\thelam) \ge 0$ ($\lambda$ has lower pre-deployment DM loss) and $Q^\F(\thelam[\mu],\thelam) \le 0$ ($\mu$ has lower pre-deployment fairness). Additionally, $\mu$ post-deployment dominates $\lambda$ if and only if:
\begin{align}
   \underbrace{\delta\dm(\thelam) -\delta\dm(\thelam[\mu])}_{\text{distributional  effect}}
   \ge
\underbrace{Q\dm(\thelam[\mu],\thelam)}_{\text{static trade-off}}
\text{ and }
\underbrace{
\delta^\F(\thelam) -\delta^\F(\thelam[\mu])
}_{\text{distributional effect}}
\ge
\underbrace{Q^\F(\thelam[\mu],\thelam) }_{\text{static trade-off}}
\end{align}
with at least one strict inequality.
\end{proposition}

This proposition shows that, after deployment, a fairer policy is at least as good on both objectives and strictly better on one when the decisions' effects offset its higher DM loss without erasing its fairness advantage. More specifically, from the first condition, as $Q\dm(\thelam[\mu], \thelam) \ge 0$, the distributional effect of the fair policy $\thelam[\mu]$ must overcome the sub-optimality in the DM objective at the pre-deployment distribution. In contrast, the second condition requires that the fair policy $\thelam[\mu]$ must exhibit an erosion in fairness lower than that of $\thelam$. Yet this proposition does not say when this happens, which we address in the next section with concrete dynamics.

\section{The Suboptimality of Unfairness}
\label{sec:results}

In this section, we specialize \cref{prop:dominance_general} to a selection process previously studied by \citet{liu2018delayed} in a lending application, whose terminology we keep; however, it can model multiple dynamic decision-making settings. 

\subsection{Selection Dynamics}

\begin{assumption}[Selection dynamics, adapted from \citet{liu2018delayed}]
\label{as:selective_dyn}
A DS from group $g$ and pre-deployment score $x$ (with density $p^g(x)$) has their score updated after decision according to:
\begin{align}
    X_\theta \gets \begin{cases}
        X + \bp &\text{ if  } \pi_\theta(X,G) = 1  \text{ and }  Y= 1  \text{ (acceptance, repayment)}\\
        X + \bm &\text{ if  } \pi_\theta(X, G) = 1 \text{ and } Y = 0 \text{ (acceptance, default)} \\
        X +r  &\text{ if  } \pi_\theta(X, G) = 0  \text{ (rejection)}
    \end{cases}
\end{align}
where $\bp \in \R_{\ge0}$, $\bm \in \R_{\le 0}$ and $r \in \R_{\le 0}$. A new label is sampled from the updated score $Y_\theta \mid X_\theta, G \sim \mathrm{Bern}(\alpha(X_\theta, G))$. Group rates $\rho^g := P(G =g) > 0$ and the conditional label distribution $\alpha(\cdot, \cdot)$ of $Y \mid X, G$ do not depend on $\theta$. 
\end{assumption}

\paragraph{Examples} 
The selection dynamics can model a range of real-world processes. In lending, $\bp > 0$ is the increase in credit score after a successful repayment, $\bm < 0$ is the decrease after a default, and we set $r=0$, so rejections leave the score unchanged. For a DS with a low chance of repayment, acceptance can hurt relative to rejection, since acceptance would likely lead to default and a lower score. In medical settings, $x$ represents a prognosis score (e.g., reflecting disease stage), and $y = 1$ represents a positive response to a treatment. Here we may set $\bp > 0$, $\bm = 0$, and $r < 0$: an ineffective treatment does not worsen the patient's condition, whereas declining treatment allows it to deteriorate.

We assume, without loss of generality, that $\upre[{\thelam[0]}]{1} > \upre[{\thelam[0]}]{0}$ (i.e., group $1$ has higher pre-deployment DS loss with the policy $\thelam[0]$ that minimizes the pre-deployment DM loss), and call group $1$ the \textit{disadvantaged group}, and the other group as \textit{advantaged}. 

\subsection{The Societal Benefit of Fairness}

Our analysis focuses on the effect for each group of increasing fairness weight $\lambda$, from its effect on the pre-deployment PO policy $\thelam$ to its distributional effects. All following results depend on \cref{as:outcome_model,as:selective_dyn}, and we state only the additional conditions. We state our main result informally, with a formal presentation in \cref{app:final_theo}.

{
\centering
\begin{tcolorbox}[
    width=\textwidth,
    colback=lightgrey,        
    colframe=white,        
    arc=3mm,                  
    boxrule=1pt,            
    boxsep=1em,               
    left=0pt, right=0pt, top=0pt, bottom=0pt 
]

\begin{theorem}[The societal benefit of fair decision-making]
\label{theo:main}
Suppose fairness is Rawlsian, the DS cost is $\Brej$ (or $\Bfn$ when rejections do not have a worse effect on qualification than a default), and accepting a DS of the disadvantaged group at the DM's break-even point $\overline c$ leaves them, on average, in a state that is cheaper for the DM than rejecting them. Then, there exists $\lambda^\dagger > 0$ such that every $\lambda \in [0, \lambda^\dagger)$ is post-deployment dominated by some $\mu \ge \lambda^\dagger$. Under egalitarian fairness, the same holds if the decrease in post-deployment DM loss within the disadvantaged group outweighs the increase in loss within the advantaged group, and there is no fairness overshoot (the disadvantaged group does not have lower post-deployment DS loss than the other group).
\end{theorem}

\end{tcolorbox}
}
In other words, under these conditions, ignoring fairness, or giving it only a small weight, is never the best choice: some fairer policy in the pre-deployment PO set has, once deployed, a DM loss no higher and a fairness loss no worse, with at least one strictly better.

We can understand this result by analyzing policies with thresholds near the pre-deployment DM optimal policy $\thelam[0]$, which accepts DS whenever their qualification exceeds the break-even point $\overline c$. Accepting DS that have qualification $\alpha(x, g) = \overline c$, or slightly below, has low costs for the DM. After deployment, however, these applicants can increase their qualification upon repayment. If this change lowers the DM's future costs, the pre-deployment trade-off between DM loss and fairness does not persist after deployment. \cref{fig:results} illustrates this argument, showing the decision effects for a DS from the disadvantaged group, with dynamics such that repayment increases the score ($\bp > 0$), default decreases it ($\bm < 0$), and rejection keeps it fixed ($r = 0$):

\begin{figure}[t]
    \centering
    \includegraphics[width=0.9\linewidth]{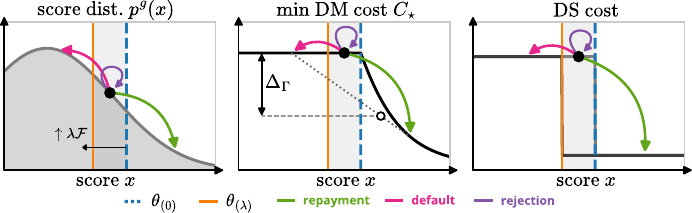}
    \caption{Changes (indicated by arrows) of a DS from the disadvantaged group (marked by the black dot $\bullet$) that is rejected by the pre-deployment DM optimal policy $\thelam[0]$ but accepted by another policy $\thelam$ with $\lambda > 0$. Their score updates to a new value based on acceptance/rejection and repayment/default. \textbf{Left:} increasing $\lambda$ moves the threshold from $\thelam[0]$ to $\thelam$, accepting the DS in the shaded band. \textbf{Center:} the DM's cost of the best decision for any DS $(x, g)$, which is independent of the marginal distribution of scores. A DS in the band who repays moves to a higher score with less risk for the DM; $\Delta_\Gamma$ is the DM's expected gain from accepting rather than rejecting. \textbf{Right:} the DS cost, which is non-null with rejection. A DS that repays increases their score and remains accepted, while one that defaults drops below the threshold $\thelam$ and gets rejected, as it already was under $\thelam[0]$.}
    \label{fig:results}
\end{figure}

\textit{Step 1: fairness increases the acceptance of the disadvantaged group} (\cref{fig:results}--left panel, \cref{sec:fairness_weight_accept_rate}). For both fairness notions, increasing $\lambda$ never raises the disadvantaged group's threshold. The threshold moves from $\thelam[0]$ (dashed) to $\thelam$ (solid), and DS in the shaded band, whose qualification is just below $\overline c$, are now accepted.

\textit{Step 2: newly accepted DS will present lower cost for the DM} (\cref{fig:results}--center panel, \cref{sec:accept_rate_dm_loss}). The panel shows the DM's cost of the best decision for each score, given by $C_\star(x, g):= \min \{C_0(x, g), C_1(x, g) \}$. If a DS in the band is rejected, their score does not change (purple). If it is accepted, their score rises after repayment (green), reaching a region where the DM's cost is lower; if they default, it falls (pink). The \textit{acceptance gap} $\Delta_\Gamma$ measures how much cheaper this DS is for the DM after acceptance (in expectation, indicated by the hollow circle) than after rejection. When $\Delta_\Gamma > 0$ at $\thelam[0]$, the condition in \cref{theo:main}, a small increase in the acceptance rate strictly lowers the post-deployment DM loss. 

\textit{Step 3: the same DS do not face higher costs, so fairness does not deteriorate} (\cref{fig:results}--right panel, \cref{sec:accept_rate_fairness}). Under $\thelam[0]$, a DS in the band is rejected and, after deployment, remains rejected because their score cannot rise. Under $\thelam$, a DS who repays remains accepted, and one who defaults may fall below $\thelam$ and be rejected. Then, decreasing the threshold never increases the post-deployment group loss of a DS that prefers acceptance, and post-deployment fairness does not worsen as $\lambda$ increases.

Combining the steps, Step 3 holds for any pair of weights $\lambda < \mu$, while Step 2 guarantees a strict DM gain for small positive weights. Together, they give the conditions of \cref{prop:dominance_general}: \emph{the distributional gains in DM loss outweigh the static suboptimality, and the fairness advantage is preserved.}

Egalitarian fairness requires the two additional conditions of \cref{theo:main} because it can also reduce unfairness by worsening the advantaged group. Step 1 then raises the advantaged group's threshold, and Step 2 runs in reverse for that group: marginal advantaged DS are now rejected, which can increase the DM loss on them. Dominance therefore requires the disadvantaged group's gain to outweigh this loss. In Step 3, the gap between groups shrinks only as long as the disadvantaged group does not overtake the advantaged one after deployment (no overshoot). Next, we present this analysis in greater detail, with proofs in \cref{app:proofs}.

\subsection{Fairness Importance vs Per-group Acceptance Rate}
\label{sec:fairness_weight_accept_rate}
Step 1 (left panel of \cref{fig:results}) argues that a positive fairness weight accepts new DS. Recall that $\theta^g$ is the threshold for group $g$, that $\thelam$ is the pre-deployment PO policy with weight $\lambda$, and that $\thelam[0]$ is the DM-optimal policy, which accepts exactly the DS with qualification $\alpha(x, g) \ge \overline c$. In this subsection, all quantities refer to pre-deployment. 

A policy sets the acceptance rate of each group, and with \cref{as:outcome_model}, this relation is bijective, so a policy $\thelam$ can be fully determined by its pre-deployment acceptance rate. Additionally, each DS group loss depends only on the respective group threshold $\theta^g$, for both $\upre{g}$ and $\upost{g}$. When DS cost matrices satisfy that acceptance never costs more than rejection, which holds for both $\Brej$ and $\Bfn$, the DS group loss is continuous and strictly increasing in the threshold $\theta^g$. Based on this property, we formalize the effect of the fairness weight $\lambda$ on thresholds $\thelam^g$. 

\begin{proposition}
\label{prop:threshold}
Let $0 \le \lambda < \mu \le 1$, let $ \C$ be any DM cost, and let the DS cost matrix be either $\Brej$ or $\Bfn$. With Rawlsian fairness $\Frw$, $\thelam[\mu]^g \le \thelam^g$ for both groups $g$. With egalitarian fairness $\Feg$, $\thelam[\mu]^{1} \le \thelam^{1}$ and $\thelam[\mu]^{0} \ge \thelam^{0}$.
\end{proposition}

Increasing the fairness weight moves the thresholds in a predictable direction. Under the Rawlsian notion, no group's acceptance rate goes down. Under the egalitarian notion, improving fairness can also mean accepting fewer DS from the advantaged group. The DS that change decisions are those between $\thelam^1$ and $\thelam[0]^1$ (shaded in \cref{fig:results}). The next two subsections examine the effect of these newly accepted DS on post-deployment DM loss and fairness.

\subsection{Per-group Acceptance Rate vs. Post-Deployment DM Loss}
\label{sec:accept_rate_dm_loss}
Step 2 (center panel) asks whether the DS in the band become cheaper for the DM after deployment. Because each DS updates independently under dynamics from \cref{as:selective_dyn}, the newly accepted DS from this band are also the only DSs whose post-deployment scores differ between $\D(\thelam[0])$ and $\D(\theta)$. Recall that $\C_\star(x, g)$ is the best cost a DM can face for DS $(x, g)$ and by \cref{prop:optimal_dm}, the pre-deployment optimum $\thelam[0]$ always selects the decision with lower cost. However, DS will alter its minimal cost through the effect of decisions. We call the \textit{expected minimal cost after deployment}, given a decision $d$, as $\Gamma_d(x,g)$ and the \textit{acceptance gap} $\Delta_\Gamma(x, g) := \Gamma_0(x, g) - \Gamma_1(x, g)$. We have that:
\begin{align}
\label{eq:gamma}
    \Gamma_1(x,g) := \underbrace{\alpha(x,g)\C_\star(x+ b^+, g)}_{\text{repays}} + \underbrace{(1 - \alpha(x,g))\C_\star(x+b^-, g)}_{\text{defaults}} \qquad \Gamma_0(x, g) := C_\star\dm(x+r, g)
\end{align}
If $\Delta_\Gamma(x, g) > 0$, accepting the DS leaves them in a state that is cheaper for the DM than rejecting it.

\begin{proposition}
\label{prop:threshold_effect_dm}
Write $\alpha^+:= \alpha \left (\thelam[0]^g+ b^+, g \right )$, $\alpha^-:= \alpha \left (\thelam[0]^g+ b^-, g \right )$ and $\alpha^r:= \alpha \left (\thelam[0]^g+ r, g \right )$ as the qualification of a DS at the break-even point after repayment ($+ \bp)$, after default ($+\bm$) and after rejection ($+ r$), respectively. For any group $g$, the post-deployment DM loss of group $g$ is differentiable w.r.t. $\theta^g$ at $\thelam[0]^g$ and the derivative has value:
\begin{align}
    \dfrac{\partial}{\partial \theta^g}\lpostg[\theta^g]{g} \Big |_{\theta^g = \thelam[0]^g} &= \Delta_\Gamma\left (\thelam[0]^g, g \right ) p^g \left (\thelam[0]^g \right ) \\
    \Delta_\Gamma \left (\thelam[0]^g, g \right ) &= \begin{cases} c_{FP}(\alpha^+ - \overline c)  & \text{ if }  \Csel \\
    c_{FP}\overline c(\alpha^+ - \overline c) + c_{FN}(1 - \overline c)(\overline c - \alpha^-) - c_{FN}(\overline c - \alpha ^r) & \text{ if } \Cacc
    \end{cases}
\end{align}
If $\Delta_\Gamma (\thelam[0]^g, g ) > 0$, a small decrease in the threshold $\thelam[0]^g$ strictly reduces the post-deployment DM loss for group $g$, and increasing the threshold strictly increases it. The opposite holds for $\Delta_\Gamma(\thelam[0]^g, g) < 0$.
\end{proposition}

The DM can obtain a better post-deployment loss by increasing the acceptance rate --- and accepting DS slightly below the break-even level $\overline c$ --- when acceptance at the margin $\overline c$ updates the score of DS to a value that costs less than rejection. In our lending example, this corresponds to DS increasing their qualification after a first approved loan. The rate of the gain grows with the number of DS at the margin. This follows from writing the loss of a policy $\theta$ as the minimal cost $\C_\star(x, g)$ plus the additional cost paid when $\theta$ disagrees with $\thelam[0]$ over the induced distribution $\D(\theta)$.

Selective and accuracy costs have different expressions for $\Delta_\Gamma (\thelam[0]^g, g  )$, which lets us interpret when increasing the acceptance rate is beneficial. With selective costs, a slightly higher acceptance rate benefits the DM when repayment positively affects DS qualification (i.e., when $\bp > 0$); under accuracy costs, however, the DM can benefit from default. For the accuracy costs, both terms $(\overline c - \alpha^-)$ and $(\overline c - \alpha^r)$ are non-negative, with the first reflecting the decrease in qualification after a default and the second the decrease in qualification after a rejection. For example, if $ r=0$, a DM improves its accuracy costs both by the increase in qualification from repayments and by the decrease from defaults. Therefore, a post-deployment DM gain from lowering a threshold is not evidence that the accepted DS are better off, which we study in the next subsection.

\subsection{Per-group Acceptance Rate vs. Post-Deployment Fairness}
\label{sec:accept_rate_fairness}

Step 3 (right panel \cref{fig:results}) discusses whether the DM gain from Step 2 comes at the expense of post-deployment fairness. Under \cref{as:selective_dyn}, re-applying $\theta$ after score updates changes only one kind of decision. A rejected DS will not increase its qualification $x + r \le x < \theta^g$ and remains rejected. An accepted DS that repays moves to $x + \bp \ge x \ge \theta^g$ and remains accepted. The only DS whose decision changes is an accepted DS that defaults and falls below the threshold, $x + \bm < \theta^g$, i.e., the defaulter DS with $x \in [\theta^g, \theta^g - \bm)$. 

In Step 2, we tracked the DM's cost through $\Gamma_d$ and $\Delta_\Gamma$; here, we use a similar approach to assess the DS's post-deployment cost. Let $\beta(x, g)$ be the expected cost a DS faces if rejected (averaging over the randomness of $Y$), which has value $\beta(x,g) = \dsc$ with rejection costs $\Brej$ and $\beta(x,g) = \dsc \alpha(x, g)$ with false negative costs $\Bfn$. After deployment and decision effects, the DS faces costs only if rejected, which we write as $\kappa_d(x, g)$, the \textit{expected post-deployment DS cost in the case of rejection} given a decision $d$. $\kappa_0$ reflects the cost of a rejected DS that remains rejected, and $\kappa_1$ reflects the cost of a DS that was accepted but gets rejected after decision effects, possible only if $\theta^g \le x \le \theta^g - \bm$. Then, the \textit{DS acceptance gap} $\Delta_\kappa(x, g) := \kappa_0(x, g) - \kappa_1(x, g)$. We have:
\begin{align}
\label{eq:kappa}
    \kappa_1(x, g) := (1 - \alpha(x, g))\,\beta(x + \bm, g) \qquad \kappa_0(x, g) := \beta(x + r, g)
\end{align}
If $\Delta_\kappa(x, g) > 0$, accepting the DS leaves them with a lower expected post-deployment cost than rejecting them. We use these ideas to prove the effect of the fairness weight on post-deployment fairness.
\begin{proposition}
\label{prop:post_fairness}
Let the DS cost be $\Brej$ or $\Bfn$. Then $\Delta_\kappa(x, g) \ge 0$ for all $(x, g)$ with $\Brej$ and, with $\Bfn$, if $r \ge \bm$. Also:
\begin{enumerate}
    \item If $\phi^g \le \theta^g$, i.e., $\phi$ accepts at least as many DS of group $g$ as $\theta$, and $\Delta_\kappa(x, g) \ge 0$ for all $x \in [\phi^g, \theta^g]$, then $\upost[\phi]{g} \le \upost{g}$.
    \item Let $0 \le \lambda < \mu \le 1$ and let $\Delta_\kappa(\cdot, g) \ge 0$ on the interval between $\thelam[\mu]^g$ and $\thelam^g$ for both groups. With Rawlsian fairness, $\frwpost[{\thelam[\mu]}] \le \frwpost[{\thelam[\lambda]}]$, and if $\upost[{\thelam[\mu]}]{1} \ge \upost[{\thelam[\mu]}]{0}$, the same holds for egalitarian fairness.
\end{enumerate}
\end{proposition}

This result first shows that increasing the acceptance rate of a group --- by decreasing the threshold --- will never make its post-deployment loss worse when the newly accepted DS have a non-negative DS acceptance gap. With $\Brej$, every rejection costs the same, so this always holds. With $\Bfn$, the sign of the gap depends on how much rejections and defaults lower qualification: the condition controls how harshly rejections can affect post-deployment qualification, and it always holds if rejections do not lower the score more than defaults ($r \ge \bm$). When the condition fails, raising the threshold can reduce post-deployment $\Bfn$ loss because the newly rejected DS become less qualified and stop counting as false negatives. This decrease does not reflect a better outcome for the group, since $\Bfn$ counts only rejected DS that are still qualified. In this regime, the $\Bfn$ fairness loss rewards policies that harm DS, which we discuss later.

The second result connects to the fact that increasing the fairness weight $\lambda$ increases the acceptance rate of the disadvantaged group, combined with the condition on the DS acceptance gap $\Delta_\kappa$, to show that a higher fairness weight does not worsen post-deployment fairness. With egalitarian fairness, this holds if the disadvantaged group still has a worse loss after the dynamics. The condition relates to overshooting: when the fairer policy helps the disadvantaged group so much that, after deployment, it becomes the better-off group, the loss gap can increase in the opposite direction.

One important aspect, previously discussed by \citet{liu2018delayed}, is whether improvements in the fairness measure benefit the DS. In our results, we compare post-deployment fairness between policies $\thelam$ and $\thelam[\mu]$; however, we still need to analyze whether a group's average condition improves. After acceptance, the expected change in the score of a DS $(x, g)$ is $\Delta_x:= \alpha(x, g) (x + \bp)  + (1 - \alpha(x, g)) (x + \bm)   - x$. This change is negative when $\alpha(x, g) < (-\bm)/(\bp - \bm)$, i.e., accepting a DS with qualification below this threshold decreases their score and, consequently, their qualification. If $\overline c > (-\bm)/(\bp - \bm)$, the policy $\thelam[0]$ will reject such applicants. Then, slightly increasing the fairness weight $\lambda > 0$ can produce thresholds $\thelam$ such that $\overline c > \alpha(\thelam^g, g) > (-\bm)/(\bp - \bm)$, resulting in lower post-deployment fairness without decreasing qualification.

\begin{figure}[t]
    \centering
    \includegraphics[width=0.95\linewidth]{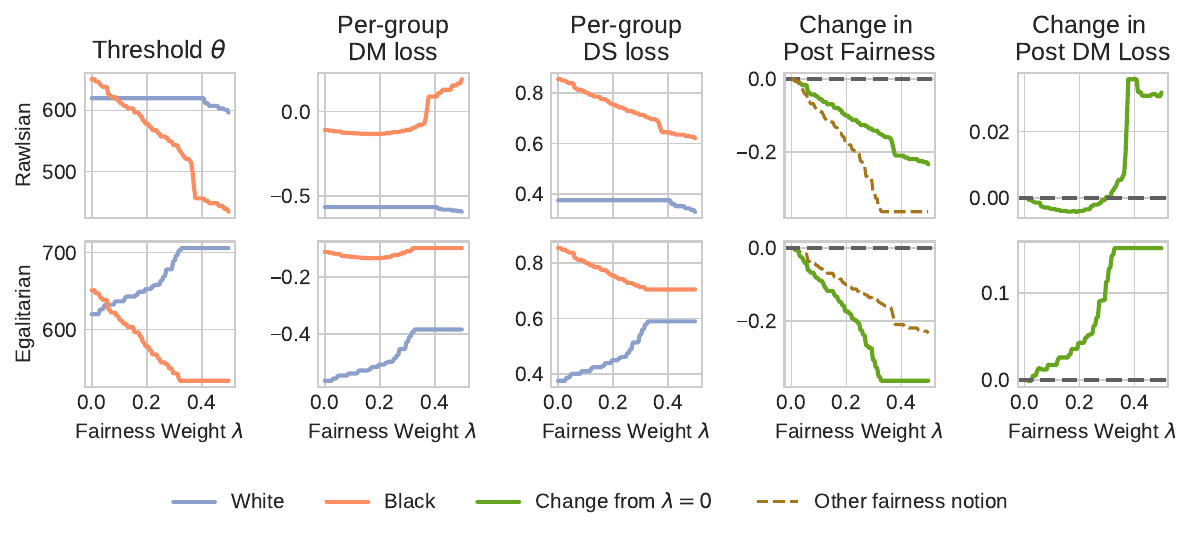}
    \caption{Effects of fairness weight on threshold and losses with selective costs for the DM and rejection costs for the DS, under Rawlsian fairness (first row) and egalitarian fairness (second row). The last two columns show the change in post-deployment losses relative to $\thelam[0]$, with the brown dashed line displaying the other fairness notion.}
    \label{fig:cases}
\end{figure}

\section{Empirical Analysis}
\label{sec:experiments}

In this section, we extend our analysis to numerical experiments with FICO data~\citep{fico}, and evaluate 1) the differences between egalitarian and Rawlsian fairness, 2) the magnitude of effects on the population and post-deployment dominance, and 3) whether accuracy costs may present post-deployment dominance that worsens DS qualification. 

\paragraph{Data}
We use the FICO dataset~\citep{fico}, previously used to study long-term effects of fair machine learning~\citep{liu2018delayed}, to set the pre-deployment distribution $\D\pre$. The data, compiled by the US Federal Reserve, contain TransUnion TransRisk credit scores $x \in [300, 850]$ for a sample of US consumers. An individual is labeled a good risk ($y = 1$) if they did not fall 90 days or more behind on any account during the following 18 months. Following \citet{liu2018delayed}, we restrict the data to white and Black DS, who make up 88\% and 12\% of this subset, respectively. For each group $g$, we compute the empirical score distribution $p^g(x)$ and the empirical repayment rate $\alpha(x, g)$. The empirical conditional distribution does not satisfy the monotonicity condition (\cref{as:outcome_model}), but we kept it as is to evaluate the robustness of our results. Black DS have lower scores on average and face a higher rejection rate under the DM-optimal policy $\thelam[0]$, so they are the disadvantaged group.

\paragraph{Implementation}
We simulate the decision process with a pool of $500,000$ applicants sampled from $\D\pre$. For each policy $\theta$, we simulate sampling from $\D(\theta)$ by applying the dynamics (\cref{as:selective_dyn}) to each DS from this initial pool. To construct the pre-deployment PO set $\mathcal S$, we select $200$ weights $\lambda \in [0, 1]$ and find the minimizer of the scalarized objective (\cref{eq:scalarization}) by doing a line search over thresholds $\theta^g$. After collecting this set, we evaluate the DM loss and fairness of each policy $\thelam$ on its respective distribution $\D(\thelam)$. In this section, we present results with the rejection cost for the DS with $\dsc = 1$, so DS group losses are the rejection rate for each group. We describe the complete details of our optimization process for constructing the pre-deployment PO set in \cref{app:implementation}.

\begin{figure}[t]
    \centering
    \includegraphics[width=\linewidth]{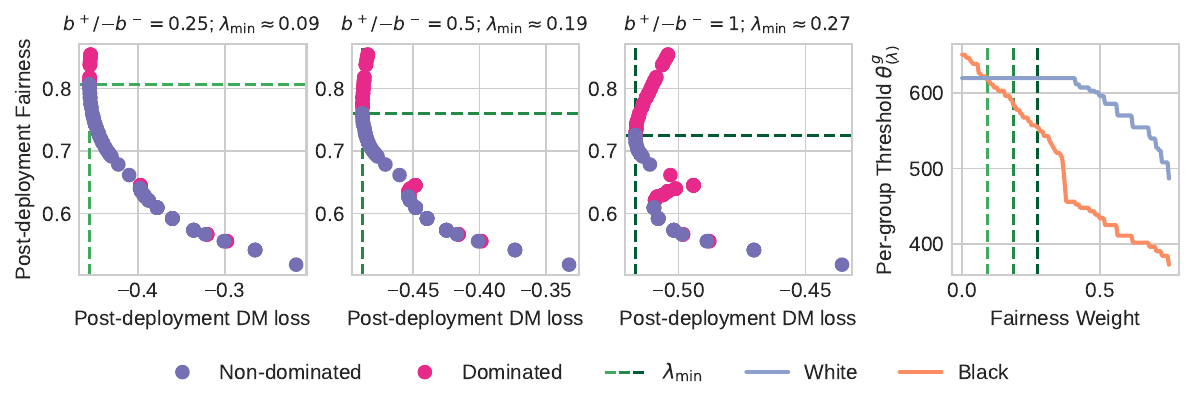}
    \caption{Post-deployment DM loss and fairness of policies from the pre-deployment PO set $\mathcal S$ for three configurations of the ratio $\bp/(-\bm)$, with selective DM costs and rejection DS costs.  Right: per-group thresholds, which do not depend on the dynamics. In all cases, fairness weights below $\lambda_{\min}$ are post-deployment dominated, and $\lambda_{\min}$ increases with the positive effect of acceptance $\bp/(-\bm)$.}
    \label{fig:improvements_effect}
\end{figure}

\subsection{Results}

\paragraph{Rawlsian vs.\ egalitarian fairness}
\cref{fig:cases} uses selective DM costs with $C_{TP} = -1, C_{FP} = 5$, and $\bp = 75$, $\bm = -150$, $r = -50$, and displays thresholds, DM loss, and fairness behavior for Rawlsian (first row) and egalitarian notions (second row). Both notions lower the threshold for Black applicants; however, for white applicants, Rawlsian fairness keeps the threshold nearly unchanged around $600$ while egalitarian fairness raises it to $700$ when $\lambda$ is close to $0.4$. As $\bp > 0$,  the acceptance gap $\Delta_\Gamma(\thelam[0]^g, g) > 0$ for both groups. For this reason, increasing the acceptance rate for Black applicants reduces post-deployment DM loss (with $\lambda$ between $0$ and $0.2$), while decreasing acceptance for white applicants increases DM loss within this group. Under egalitarian fairness, the effect of white applicants, who are the majority, dominates the positive effect of increasing the acceptance rate of Black applicants, and the lowest post-deployment loss occurs when $\lambda = 0$. With only two groups, the two fairness notions are related, and improvements in one are reflected in the other, as the dashed line shows.

\paragraph{Magnitude of Decision Effects and Post-Deployment Dominance}
We now vary how much repayment raises a score, setting $\bp \in \{37.5, 75, 150\}$ while keeping selective DM costs, Rawlsian fairness, $\bm = -150$, and $r = -50$ (\cref{fig:improvements_effect}). The ratio $\bp / (-\bm) \in \{0.25, 0.5, 1\}$ captures how much a loan can help a borrower relative to how much it can hurt them. A ratio of $0.5$, as in \citet{liu2018delayed} (who use $r = 0$), describes a setting where a default lowers a score much more than a repayment raises it. A ratio of $1$ describes a setting where repayment is rewarded more strongly (e.g., a first-time borrower program), and the ratio can also reflect broader economic conditions. Under selective costs, the acceptance gap $\Delta_\Gamma(\thelam[0]^g, g) = c_{FP}(\alpha^+ - \overline c)$ is positive whenever $\bp > 0$ (\cref{prop:threshold_effect_dm}). Lowering the threshold for Black applicants slightly below the DM-optimal one therefore reduces the post-deployment DM loss. Accordingly, in all three settings, every weight below $\lambda_{\min}$ is post-deployment dominated, with $\lambda_{\min} \approx 0.09$, $0.19$, and $0.27$, respectively. At $\lambda_{\min}$, the threshold for Black applicants drops from about $655$ to about $620$, $580$, and $550$, while the threshold for white applicants is unchanged. Thus, the larger the benefit of repayment relative to the cost of default, the larger the range of fairness weights that are post-deployment dominated, and the stronger the case for a higher fairness weight.

\begin{figure}
\centering
\includegraphics[width=0.9\linewidth]{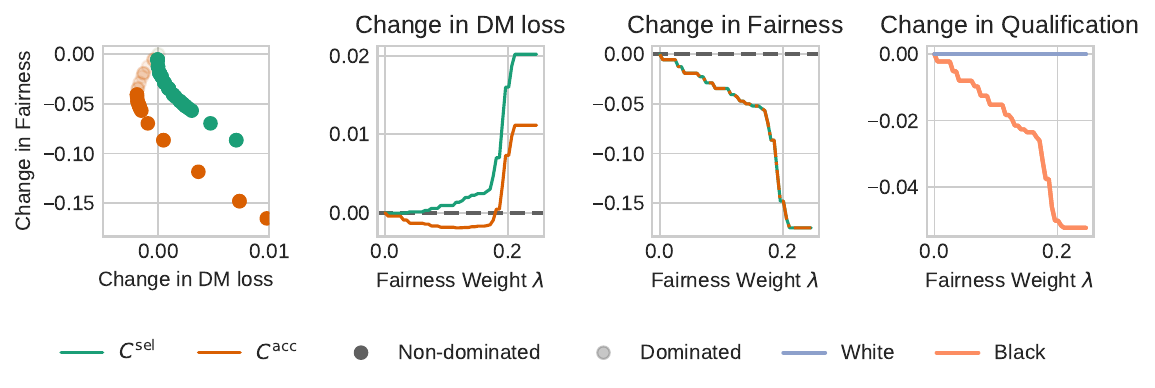}
\caption{Difference in post-deployment dominance between selective costs and accuracy costs when scores cannot increase and will only decrease with default. Values are displayed as the change in comparison to $\thelam[0]$. The two cost matrices yield the same policies, so fairness and qualification coincide; accuracy improves while qualification for Black applicants falls.}
\label{fig:acc_sel}
\end{figure}

\paragraph{Post-Deployment Accuracy Gains Can Come From Harming DS}
Finally, we test whether an accuracy-based DM loss can reward policies that make DS less qualified. We consider the extreme case $\bp = r = 0$ and $\bm = -150$, where scores can never increase and can only drop after a default. We use Rawlsian fairness with rejection costs and compare symmetric accuracy costs ($c_{FP} = c_{FN} = 1$) with symmetric selective costs ($c_{FP} = -c_{TP} = 1$). Both give $\overline c = 0.5$, so $\thelam[0]$ approves an applicant whenever repayment is more likely than default, and DS accepted by $\thelam$ with $\lambda > 0$ will more commonly default than repay. In \cref{fig:acc_sel}, both cost matrices yield the same post-deployment fairness for each $\lambda$, but their DM losses differ. As $\lambda$ increases, the post-deployment qualification of Black applicants, $\E_{\D(\thelam)}[\alpha(X, G) \mid G = 1]$, decreases. The selective loss never improves on $\thelam[0]$, so it has no post-deployment dominated policies. In contrast, accuracy loss improves as this qualification decreases, so small fairness weights are post-deployment dominated. This happens because accuracy costs also penalize rejecting applicants who would repay: once accepted applicants default and their scores drop, more of them are correctly rejected, and the DM's post-deployment loss decreases.

\section{Conclusion and Future Work}

In this work, we studied the trade-offs between performance and fairness when decisions reshape the population. Starting from a general multi-objective framework, we define the DM loss and each DS group's loss as costs over decisions and outcomes. We then construct a set of Pareto-optimal (PO) policies by varying a fairness weight, where each policy is PO on pre-deployment distribution. Our main question is whether policies from this set introduce societal costs, that is, display worse loss for one stakeholder without improving the other, in comparison to other policies from this set. Our results first show simple conditions under which this can occur for general losses and dynamics, which we later specialize to a selection dynamic.

Interesting directions for future work include extending our one-step analysis to multiple rounds and connecting to ideas of stable solutions from performative prediction. Additionally, future work involves designing algorithms that learn system dynamics without exposing DS to unfair decisions, e.g., by leveraging historical data.

\newpage

\section*{Ethical Considerations Statement}

The presented research addresses fair decision-making in social contexts, using lending as its main motivating example. Our empirical analysis uses the FICO dataset~\citep{fico}, compiled by the US Federal Reserve from TransUnion credit scores, which includes race as a sensitive attribute. We use only group-level aggregate statistics, namely the score distribution and the repayment rate at each score. We did not access individual-level records; the data contain no personally identifiable information.

This work contributes to ongoing efforts in fairness-aware decision-making by framing decision-making as a multi-stakeholder system and investigating the consequences of unfair policies when decisions reshape the population they interact with. However, unfairness and discrimination arise from social and historical processes, and technical evaluations and interventions alone cannot identify or address broader sources of bias. In particular, credit scores and repayment records may themselves reflect structural inequality, and our analysis takes them as given.

Our findings are intended to inform the evaluation of decision policies, and any application should account for the specific context. They rely on stylized population dynamics, and the policies we study apply group-specific thresholds, whose use may be legally restricted in domains such as lending. Moreover, our results should not be read as implying that fairness is justified only when it also reduces the DM's loss. Rather, they show that neglecting fairness can lead to post-deployment outcomes in which no stakeholder is better off than under a fairer policy, and at least one is worse off.

\section*{Generative AI Usage Statement}

In this work, we used generative AI tools (Claude, Opus 5.5) to assist with code development and improve the manuscript's presentation.  We did not use generative AI tools to develop the research ideas, outline sections, or interpret the experimental results; the authors developed and analyzed these. We reviewed the correctness of the generated code and the suggested text edits. We take full responsibility for the final content of this work.

\bibliography{main}

@inproceedings{corbett2017algorithmic,
	title        = {Algorithmic decision making and the cost of fairness},
	author       = {Corbett-Davies, Sam and Pierson, Emma and Feller, Avi and Goel, Sharad and Huq, Aziz},
	year         = 2017,
	booktitle    = {ACM SIGKDD International Conference on Knowledge Discovery and Data Mining},
	pages        = {797--806}
}

@inproceedings{perdomo2020performative,
  title={Performative Prediction},
  author={Perdomo, Juan and Zrnic, Tijana and Mendler-D{\"u}nner, Celestine and Hardt, Moritz},
  booktitle={International Conference on Machine Learning},
  pages={7599--7609},
  year={2020},
  organization={PMLR}
}

@inproceedings{liu2018delayed,
  title={Delayed impact of fair machine learning},
  author={Liu, Lydia T and Dean, Sarah and Rolf, Esther and Simchowitz, Max and Hardt, Moritz},
  booktitle={International Conference on Machine Learning},
  pages={3150--3158},
  year={2018},
  organization={PMLR}
}

@inproceedings{rateike2024designing,
  title={Designing long-term group fair policies in dynamical systems},
  author={Rateike, Miriam and Valera, Isabel and Forr{\'e}, Patrick},
  booktitle={Proceedings of the 2024 ACM Conference on Fairness, Accountability, and Transparency},
  pages={20--50},
  year={2024}
}

@misc{fico,
author={{US} Federal Reserve},
title={Report to the Congress on credit scoring and its effects on the availability and affordability of credit},
year={2007}
}

@article{zhang2020fair,
  title={How do fair decisions fare in long-term qualification?},
  author={Zhang, Xueru and Tu, Ruibo and Liu, Yang and Liu, Mingyan and Kjellstrom, Hedvig and Zhang, Kun and Zhang, Cheng},
  journal={Advances in Neural Information Processing Systems},
  volume={33},
  pages={18457--18469},
  year={2020}
}

@InProceedings{kilbertus2020fair,
  title = 	 {Fair Decisions Despite Imperfect Predictions},
  author =       {Kilbertus, Niki and Rodriguez, Manuel Gomez and Sch\"olkopf, Bernhard and Muandet, Krikamol and Valera, Isabel},
  booktitle = 	 {Proceedings of the Twenty Third International Conference on Artificial Intelligence and Statistics},
  pages = 	 {277--287},
  year = 	 {2020},
  editor = 	 {Chiappa, Silvia and Calandra, Roberto},
  volume = 	 {108},
  series = 	 {Proceedings of Machine Learning Research},
  month = 	 {26--28 Aug},
  publisher =    {PMLR},
}

@inproceedings{lakkaraju2017selective,
author = {Lakkaraju, Himabindu and Kleinberg, Jon and Leskovec, Jure and Ludwig, Jens and Mullainathan, Sendhil},
title = {The Selective Labels Problem: Evaluating Algorithmic Predictions in the Presence of Unobservables},
year = {2017},
isbn = {9781450348874},
publisher = {Association for Computing Machinery},
address = {New York, NY, USA},
booktitle = {Proceedings of the 23rd ACM SIGKDD International Conference on Knowledge Discovery and Data Mining},
pages = {275–284},
numpages = {10},
location = {Halifax, NS, Canada},
series = {KDD '17}
}

@inproceedings{raab2024fair,
  title={Fair participation via sequential policies},
  author={Raab, Reilly and Boczar, Ross and Fazel, Maryam and Liu, Yang},
  booktitle={Proceedings of the AAAI Conference on Artificial Intelligence},
  volume={38},
  number={13},
  pages={14758--14766},
  year={2024}
}

@inproceedings{hu2022achieving,
  title={Achieving long-term fairness in sequential decision making},
  author={Hu, Yaowei and Zhang, Lu},
  booktitle={Proceedings of the AAAI Conference on Artificial Intelligence},
  volume={36},
  number={9},
  pages={9549--9557},
  year={2022}
}

@inproceedings{wen2021algorithms,
	title        = {Algorithms for fairness in sequential decision making},
	author       = {Wen, Min and Bastani, Osbert and Topcu, Ufuk},
	year         = 2021,
	booktitle    = {International Conference on Artificial Intelligence and Statistics},
	pages        = {1144--1152},
	organization = {PMLR}
}

@article{mishler2022fair,
  title={Fair when trained, unfair when deployed: Observable fairness measures are unstable in performative prediction settings},
  author={Mishler, Alan and Dalmasso, Niccol{\`o}},
  journal={arXiv preprint arXiv:2202.05049},
  year={2022}
}

@inproceedings{jin2026addressing,
  title={Addressing polarization and unfairness in performative prediction},
  author={Jin, Kun and Xie, Tian and Liu, Yang and Zhang, Xueru},
  booktitle={Proceedings of the AAAI Conference on Artificial Intelligence},
  volume={40},
  number={27},
  pages={22408--22416},
  year={2026}
}

@article{mehrabi2021survey,
author = {Mehrabi, Ninareh and Morstatter, Fred and Saxena, Nripsuta and Lerman, Kristina and Galstyan, Aram},
title = {A Survey on Bias and Fairness in Machine Learning},
year = {2021},
issue_date = {July 2022},
publisher = {Association for Computing Machinery},
address = {New York, NY, USA},
volume = {54},
number = {6},
issn = {0360-0300},
journal = {ACM Comput. Surv.},
month = {jul},
articleno = {115},
numpages = {35}
}

@inproceedings{damour2020fairness,
  title={Fairness is not static: deeper understanding of long term fairness via simulation studies},
  author={D'Amour, Alexander and Srinivasan, Hansa and Atwood, James and Baljekar, Pallavi and Sculley, David and Halpern, Yoni},
  booktitle={Proceedings of the 2020 Conference on Fairness, Accountability, and Transparency},
  pages={525--534},
  year={2020}
}

@inproceedings{gupta2026first,
  title={First-See-Then-Design: A Multi-Stakeholder View for Optimal Performance-Fairness Trade-Offs},
  author={Gupta, Kavya and Kalampalikis, Nektarios and Heitz, Christoph and Valera, Isabel},
  booktitle={The 2026 ACM Conference on Fairness, Accountability, and Transparency},
  pages={218--248},
  year={2026}
}

@Article{jung2020fair,
 author = {Christopher Jung and Sampath Kannan and Changhwa Lee and Mallesh M. Pai and Aaron Roth and R. Vohra},
 booktitle = {ACM Conference on Economics and Computation},
 journal = {Proceedings of the 21st ACM Conference on Economics and Computation},
 title = {Fair Prediction with Endogenous Behavior},
 year = {2020}
}

@article{hardt2016equality,
  title={Equality of opportunity in supervised learning},
  author={Hardt, Moritz and Price, Eric and Srebro, Nati},
  journal={Advances in neural information processing systems},
  volume={29},
  year={2016}
}

@inproceedings{somerstep2024algorithmic,
  title={Algorithmic fairness in performative policy learning: Escaping the impossibility of group fairness},
  author={Somerstep, Seamus and Ritov, Ya'acov and Sun, Yuekai},
  booktitle={Proceedings of the 2024 ACM Conference on Fairness, Accountability, and Transparency},
  pages={616--630},
  year={2024}
}

@article{chouldechova2017fair,
  title={Fair prediction with disparate impact: A study of bias in recidivism prediction instruments},
  author={Chouldechova, Alexandra},
  journal={Big data},
  volume={5},
  number={2},
  pages={153--163},
  year={2017},
  publisher={SAGE Publications, Sage CA: Los Angeles, CA}
}

@article{yin2023long,
	title        = {Long-term fairness with unknown dynamics},
	author       = {Yin, Tongxin and Raab, Reilly and Liu, Mingyan and Liu, Yang},
	year         = 2023,
	journal      = {Advances in Neural Information Processing Systems},
	volume       = 36,
	pages        = {55110--55139}
}

@article{zhang2019group,
  title={Group retention when using machine learning in sequential decision making: the interplay between user dynamics and fairness},
  author={Zhang, Xueru and Khaliligarekani, Mohammadmahdi and Tekin, Cem and others},
  journal={Advances in neural information processing systems},
  volume={32},
  year={2019}
}

@inproceedings{yu2022policy,
	title        = {Policy Optimization with Advantage Regularization for Long-Term Fairness in Decision Systems},
	author       = {Eric Yang Yu and Zhizhen Qin and Min Kyung Lee and Sicun Gao},
	year         = 2022,
	booktitle    = {Advances in Neural Information Processing Systems},
	editor       = {Alice H. Oh and Alekh Agarwal and Danielle Belgrave and Kyunghyun Cho}
}

@article{gohar2025longterm,
title={Long-Term Fairness Inquiries and Pursuits in Machine Learning: A Survey of Notions, Methods, and Challenges},
author={Usman Gohar and Zeyu Tang and Jialu Wang and Kun Zhang and Peter Spirtes and Yang Liu and Lu Cheng},
journal={Transactions on Machine Learning Research},
issn={2835-8856},
year={2025},
note={}
}

@inproceedings{wilms2026fairness,
  title={Fairness vs Performance: Characterizing the Pareto Frontier of Algorithmic Decision Systems},
  author={Wilms, Mieke and Heitz, Christoph},
  booktitle={The 2026 ACM Conference on Fairness, Accountability, and Transparency},
  pages={2885--2907},
  year={2026}
}

@article{moscato2021benchmark,
  title={A benchmark of machine learning approaches for credit score prediction},
  author={Moscato, Vincenzo and Picariello, Antonio and Sperl{\'\i}, Giancarlo},
  journal={Expert Systems with Applications},
  volume={165},
  pages={113986},
  year={2021},
  publisher={Elsevier}
}

@inproceedings{dwork2012fairness,
  title={Fairness through awareness},
  author={Dwork, Cynthia and Hardt, Moritz and Pitassi, Toniann and Reingold, Omer and Zemel, Richard},
  booktitle={Proceedings of the 3rd innovations in theoretical computer science conference},
  pages={214--226},
  year={2012}
}

@inproceedings{zafar2017fairness,
  title={Fairness constraints: Mechanisms for fair classification},
  author={Zafar, Muhammad Bilal and Valera, Isabel and Rogriguez, Manuel Gomez and Gummadi, Krishna P},
  booktitle={Artificial intelligence and statistics},
  pages={962--970},
  year={2017},
  organization={PMLR}
}

@inproceedings{zemel2013learning,
  title={Learning fair representations},
  author={Zemel, Rich and Wu, Yu and Swersky, Kevin and Pitassi, Toni and Dwork, Cynthia},
  booktitle={International conference on machine learning},
  pages={325--333},
  year={2013},
  organization={PMLR}
}

@inproceedings{cruz2023fairgbm,
title={Fair{GBM}: Gradient Boosting with Fairness Constraints},
author={Andr{\'e} Cruz and Catarina G Bel{\'e}m and Jo{\~a}o Bravo and Pedro Saleiro and Pedro Bizarro},
booktitle={The Eleventh International Conference on Learning Representations },
year={2023},
}

@inproceedings{bowman1976relationship,
  title={On the relationship of the Tchebycheff norm and the efficient frontier of multiple-criteria objectives},
  author={Bowman Jr, V Joseph},
  booktitle={Multiple Criteria Decision Making: Proceedings of a Conference Jouy-en-Josas, France May 21--23, 1975},
  pages={76--86},
  year={1976},
  organization={Springer}
}

@inproceedings{menon2018cost,
  title={The cost of fairness in binary classification},
  author={Menon, Aditya Krishna and Williamson, Robert C},
  booktitle={Conference on Fairness, accountability and transparency},
  pages={107--118},
  year={2018},
  organization={PMLR}
}

@article{wei2022fairness,
  title={The fairness-accuracy Pareto front},
  author={Wei, Susan and Niethammer, Marc},
  journal={Statistical Analysis and Data Mining: The ASA Data Science Journal},
  volume={15},
  number={3},
  pages={287--302},
  year={2022},
  publisher={Wiley Online Library}
}

@article{zhao2022inherent,
  title={Inherent tradeoffs in learning fair representations},
  author={Zhao, Han and Gordon, Geoffrey J},
  journal={Journal of Machine Learning Research},
  volume={23},
  number={57},
  pages={1--26},
  year={2022}
}

@article{wick2019unlocking,
  title={Unlocking fairness: a trade-off revisited},
  author={Wick, Michael and Tristan, Jean-Baptiste and others},
  journal={Advances in neural information processing systems},
  volume={32},
  year={2019}
}

@inproceedings{dutta2020there,
  title={Is there a trade-off between fairness and accuracy? a perspective using mismatched hypothesis testing},
  author={Dutta, Sanghamitra and Wei, Dennis and Yueksel, Hazar and Chen, Pin-Yu and Liu, Sijia and Varshney, Kush},
  booktitle={International conference on machine learning},
  pages={2803--2813},
  year={2020},
  organization={PMLR}
}

@inproceedings{kleinberg2017inherent,
  author       = {Jon M. Kleinberg and
                  Sendhil Mullainathan and
                  Manish Raghavan},
  editor       = {Christos H. Papadimitriou},
  title        = {Inherent Trade-Offs in the Fair Determination of Risk Scores},
  booktitle    = {8th Innovations in Theoretical Computer Science Conference, {ITCS}
                  2017, Berkeley, CA, USA, January 9-11, 2017},
  series       = {LIPIcs},
  volume       = {67},
  pages        = {43:1--43:23},
  publisher    = {Schloss Dagstuhl - Leibniz-Zentrum f{\"{u}}r Informatik},
  year         = {2017},
  bibsource    = {dblp computer science bibliography, https://dblp.org}
}

@book{rawls1971theory,
  address = {Cambridge, Mass},
  author = {Rawls, John},
  note = {Eleventh printing, 1981},
  publisher = {The Belknap press of Harvard University Press},
  title = {A Theory of Justice},
  year = 1971
}

@article{arneson1999egalitarianism,
author={Arneson, Richard J.},
title={Egalitarianism and Responsibility},
journal={The Journal of Ethics},
year={1999},
month={Sep},
day={01},
volume={3},
number={3},
pages={225-247},
issn={1572-8609},
}
\bibliographystyle{tmlr}

\appendix

\section{Proofs}
\label{app:proofs}

In this section, we prove all results in Sec.~\ref{sec:results}.

\subsection{Proof of \texorpdfstring{\cref{prop:dominance_general}}{Proposition \getrefnumber{prop:dominance_general}}}

\label{app:proof_dominance_general}

We first introduce an auxiliary lemma.

\begin{lemma} 
\label{lemma:solution_scalarized}
Consider two functions $f, s : \mathbb R \to \mathbb R$ and a weight $\lambda \in [0, 1]$ and define $x_\lambda \in \argmin (1 - \lambda) f(x) + \lambda s(x)$. Then, for two weights $0 \le \lambda < \mu \le 1$, $s(x_\mu) \le s(x_\lambda)$ and $f(x_\lambda) \le f(x_\mu)$.

\end{lemma}

\begin{proof}
By the optimality of $x_\lambda$, we have that:
\begin{align}
    (1 - \lambda) f(x_\lambda ) + \lambda s(x_\lambda) \le (1 - \lambda) f(x_\mu) + \lambda s(x_\mu) \implies      (1 - \lambda ) (f(x_\lambda) - f(x_\mu)) \le   \lambda (s(x_\mu) - s(x_\lambda))
\end{align}
and by the optimality of $x_\mu$ we can also conclude that $\mu(s(x_\mu) - s(x_\lambda)) \le (1 - \mu) (f(x_\lambda) - f(x_\mu))$. By summing these two inequalities, we have:
\begin{align}
    (1 - \lambda ) (f(x_\lambda) - f(x_\mu))  + \mu(s(x_\mu) - s(x_\lambda)) &\le   \lambda (s(x_\mu) - s(x_\lambda)) + (1 - \mu) (f(x_\lambda) - f(x_\mu)) \implies \\
    (\mu - \lambda ) (f(x_\lambda) - f(x_\mu))  &\le   (\lambda - \mu) (s(x_\mu) - s(x_\lambda)) \implies \\
    (\mu - \lambda ) (f(x_\lambda) - f(x_\mu) + s(x_\mu) - s(x_\lambda))   &\le 0
\end{align}
As $\mu - \lambda > 0$, we must have that $(f(x_\lambda) - f(x_\mu) + s(x_\mu) - s(x_\lambda)) \le 0$, which implies that $f(x_\lambda) - f(x_\mu)   \le s(x_\lambda) - s(x_\mu)$. We can apply this inequality again to the optimality conditions. First, for $\lambda$:
\begin{align}
    (1 - \lambda ) (f(x_\lambda) - f(x_\mu)) &\le   \lambda (s(x_\mu) - s(x_\lambda)) \le \lambda (f(x_\mu)  - f(x_\lambda)) \implies \\
    (1 - \lambda + \lambda) (f(x_\lambda) - f(x_\mu)) &\le 0 \implies  f(x_\lambda)  \le f(x_\mu)
\end{align}
For $\mu$:
\begin{align}
    \mu(s(x_\mu) - s(x_\lambda)) &\le (1 - \mu) (f(x_\lambda) - f(x_\mu)) \le (1 - \mu)(s(x_\lambda) - s(x_\mu)) \implies  
    s(x_\mu)  \le  s(x_\lambda)
\end{align}
\end{proof}

Now, we can conclude that $Q\dm(\thelam[\mu], \thelam) \ge 0$ and $Q^\F(\thelam[\mu], \thelam) \le 0$ by using \cref{lemma:solution_scalarized} with $f(x) := \lpre[\theta]$ and $s(x) := \fpre[\theta]$ and using the fact that $\lambda < \mu$.  Then we can prove the inequalities in a few simple steps. For the DM loss:
\begin{align}
    \delta\dm(\thelam) -\delta\dm(\thelam[\mu]) &\ge Q\dm(\thelam[\mu],\thelam) \Longleftrightarrow \\
    \lpost[\thelam] - \lpre[\thelam] -\lpost[{\thelam[\mu]}] +  \lpre[{\thelam[\mu]}] & \ge \lpre[{\thelam[\mu]}] - \lpre[{\thelam}] \Longleftrightarrow \\ 
    \lpost[\thelam] -\lpost[{\thelam[\mu]}]  & \ge 0  \\ 
\end{align}
The same steps apply to fairness to show that the condition implies $\fpost[\thelam] - \fpost[{\thelam[\mu]}] \ge 0$. If at least one of these inequalities is strict, then $\mu$ post-deployment dominates $\lambda$. \qed

\subsection{Proof of \texorpdfstring{\cref{prop:threshold}}{Proposition \getrefnumber{prop:threshold}}}

Importantly, we keep the conditions from \cref{prop:optimal_dm}. We will write $\lpreg[\theta^g]{g}$ for the pre-deployment DM loss evaluated only on group $g$. Also recall the expected cost of rejection $\beta(x, g)$, so that $\upre{g} = \int_{-\infty}^{\theta^g} \beta(x, g) p^g(x)\, dx$. To treat both DS costs at once, we write $\beta(x, g) = \dsc_{TN}(1 - \alpha(x, g)) + \dsc_{FN}\,\alpha(x, g)$, with $\dsc_{TN} = \dsc_{FN} = \dsc$ for $\Brej$ and $\dsc_{TN} = 0$, $\dsc_{FN} = \dsc$ for $\Bfn$. We introduce an auxiliary lemma showing that the DM loss can be written as a function of the group-wise DS losses and is monotonic on two intervals of group losses.

\begin{lemma}
\label{lemma:loss_coordinates}
For each group $g$, the map $\theta^g \mapsto \upre{g}$ is continuous and strictly increasing; denote its range by $I^g$ and its inverse by $\tau^g : I^g \to \X$. Then $\ell^g := \mathcal L_{\mathrm{pre}}^g \circ \tau^g$ is strictly decreasing on $I^g \cap (-\infty, \upre[{\thelam[0]}]{g}]$ and strictly increasing on $I^g \cap [\upre[{\thelam[0]}]{g}, \infty)$. Moreover, $\ell^g$ is strictly convex for $\Brej$, and for $\Bfn$ whenever $\overline c > 0$.
\end{lemma}

\begin{proof}
Since $\beta(\cdot, g) > 0$ and $p^g > 0$, $\upre{g}$ is strictly increasing and continuously differentiable, with derivative $\beta(\theta^g, g) p^g(\theta^g)$. As $\lpreg[\theta^g]{g} = \int_{-\infty}^{\theta^g} \C_0(x, g) p^g(x)\, dx + \int_{\theta^g}^{\infty} \C_1(x, g) p^g(x)\, dx$, we have from \cref{eq:decision_costs} that
\begin{align}
    \frac{d}{d\theta^g} \lpreg[\theta^g]{g} = \left(\C_0(\theta^g, g) - \C_1(\theta^g, g)\right) p^g(\theta^g) = \eta \left(\alpha(\theta^g, g) - \overline c\right) p^g(\theta^g).
\end{align}
By the chain rule, for $u \in I^g$ and $x := \tau^g(u)$,
\begin{align}
    (\ell^g)'(u) = \frac{\eta \left(\alpha(x, g) - \overline c\right)}{\beta(x, g)} = \eta\, h(\alpha(x, g)), \qquad h(a) := \frac{a - \overline c}{\dsc_{TN}(1 - a) + \dsc_{FN}\, a}.
\end{align}
We conclude the monotonicity of $\ell^g$ from the sign of $\alpha(x, g) - \overline c$. By \cref{prop:optimal_dm}, if $x < \thelam[0]^g$, then $\alpha(x, g) < \overline c$, and the opposite holds if $x > \thelam[0]^g$. For the convexity, the derivative of $h$ is:
\begin{align}
    h'(a) = \frac{\dsc_{TN}(1 - \overline c) + \dsc_{FN}\, \overline c}{\left(\dsc_{TN}(1 - a) + \dsc_{FN}\, a\right)^2},
\end{align}
whose numerator is positive for $\Brej$, as $\dsc_{TN}, \dsc_{FN} > 0$ and $\overline c \in [0, 1]$, and for $\Bfn$ if $\overline c > 0$. As $h$, $\alpha(\cdot, g)$ and $\tau^g$ are strictly increasing, the same holds for $(\ell^g)'$, and then $\ell^g$ is strictly convex.
\end{proof}

By \cref{lemma:loss_coordinates}, a policy is determined by its vector of group losses $v = (v^0, v^1) \in I^0 \times I^1$, and we can write the pre-deployment objectives:
\begin{align}
    \Lo(v) := \sum_{g} \rho^g \ell^g(v^g), \qquad \Frw(v) = \max\{v^0, v^1\}, \qquad \Feg(v) = |v^1 - v^0|.
\end{align}
We write $u^g(\lambda) := \upre[\thelam]{g}$ and $u(\lambda) := (u^0(\lambda), u^1(\lambda))$. Since $\tau^g$ is strictly increasing, it suffices to prove the statements for $u^g$ instead of $\thelam^g$. We use the following consequence of optimality of $\thelam$: 
\begin{align}
\label{eq:opt_consequence}
    \F(v) \le \F(u(\lambda)) \implies \Lo(v) \ge \Lo(u(\lambda)).
\end{align}
We prove the result separately for the Rawlsian and egalitarian cases. 

\paragraph{Rawlsian}
We first prove by contradiction that for every $\lambda$, $u^g(\lambda) \le u^g(0)$ (with $\lambda = 0$ we have the highest loss).
Suppose $u^g(\lambda) > u^g(0)$ for some $\lambda$ and $g$, and let $v$ be such that $v^h = u^h(\lambda)$ and $v^g = u^g(0)$. Then $\Frw(v) \le \Frw(u(\lambda))$, while $\Lo(v) < \Lo(u(\lambda))$ because $\ell^g$ is strictly increasing on $[u^g(0), u^g(\lambda)]$. This contradicts \cref{eq:opt_consequence}. Hence $u^g(\lambda) \le u^g(0)$ for all $\lambda$ and $g$.

Now, assume by contradiction that $u^g(\mu) > u^g(\lambda)$ for a group $g$, and let $v^g := u^g(\mu)$ and $v^h := u^h(\lambda)$. By \cref{lemma:solution_scalarized} with $f := \lpre[\cdot]$ and $s := \frwpre[\cdot]$ and by the definition of $\Frw$, we have $u^g(\mu) \le \Frw(u(\mu)) \le \Frw(u(\lambda))$, and $u^h(\lambda) \le \Frw(u(\lambda))$. Therefore $\Frw(v) = \max\{u^g(\mu), u^h(\lambda)\} \le \Frw(u(\lambda))$. For the DM loss,
\begin{align}
    \Lo(v) - \Lo(u(\lambda)) = \rho^g \left( \ell^g(u^g(\mu)) - \ell^g(u^g(\lambda)) \right) < 0,
\end{align}
since $u^g(\lambda) < u^g(\mu) \le u^g(0)$ (as we showed $u^g(0)$ is the highest loss) and $\ell^g$ is strictly decreasing on this interval. This contradicts \cref{eq:opt_consequence}, so $u^g(\mu) \le u^g(\lambda)$ for both groups.

\paragraph{Egalitarian}
Step 1: We first show that $u^1(\lambda) \ge u^0(\lambda)$ for all $\lambda$, i.e., the disadvantaged group remains with the highest pre-deployment loss. This holds for $\lambda = 0$ by the definition of the disadvantaged group. Suppose, by contradiction, $\lambda > 0$ and $u^1(\lambda) < u^0(\lambda)$, and let $v(t) := (1 - t)u(\lambda) + t u(0) \in I^0 \times I^1$ for $t \in [0, 1]$. The gap $v^1(t) - v^0(t)$ is continuous, negative at $t = 0$ and non-negative at $t = 1$, so it vanishes at some $t^\star \in (0, 1]$. Then $\Feg(v(t^\star)) = 0 < \Feg(u(\lambda))$. Each $v^g(t^\star)$ lies between $u^g(\lambda)$ and $u^g(0)$, so $\ell^g(v^g(t^\star)) \le \ell^g(u^g(\lambda))$ by \cref{lemma:loss_coordinates}, and thus $\Lo(v(t^\star)) \le \Lo(u(\lambda))$. As $\lambda > 0$, $v(t^\star)$ attains a strictly lower scalarized objective than $u(\lambda)$, a contradiction.

Step 2: We show that there are no weights $\lambda, \mu$ with $u^g(\mu) > u^g(\lambda)$ for both groups. Suppose otherwise, take $0 < \delta < \frac{1}{2}\min_g \{u^g(\mu) - u^g(\lambda)\}$, and define $v := u(\lambda) + \delta (1, 1)$ and $w := u(\mu) - \delta(1, 1)$, which lie in $I^0 \times I^1$. Shifting both group-losses by the same amount does not change the gap between them, so $\Feg(v) = \Feg(u(\lambda))$ and $\Feg(w) = \Feg(u(\mu))$, and \cref{eq:opt_consequence} gives $\Lo(v) - \Lo(u(\lambda)) \ge0 $ and $\Lo(w) - \Lo(u(\mu)) \ge 0$. Summing these two inequalities:
\begin{align}
    \sum_g \rho^g \big [ \underbrace{\left( \ell^g(u^g(\lambda) + \delta) - \ell^g(u^g(\lambda)) \right)}_{\ge 0 } - \underbrace{\left( \ell^g(u^g(\mu)) - \ell^g(u^g(\mu) - \delta) \right) }_{\ge 0 } \big ] \ge 0
\end{align}
Each term is positive because $\ell^g$ is strictly convex and increasing in this interval (\cref{lemma:loss_coordinates}). However, by strict convexity, $\ell^g(z + \delta) - \ell^g(z)$ increases as $z$ increases, so the right-hand side is larger, and each term in the sum is negative, which is a contradiction. An analogous proof shows that $u^g(\mu) < u^g(\lambda)$ for both groups is also impossible.

Step 3: By \cref{lemma:solution_scalarized} with $s := \fegpre[\cdot]$, we have $\Feg(u(\mu)) \le \Feg(u(\lambda))$, which follows from the fact that the disadvantaged always has higher loss:
\begin{align}
    u^1(\mu) - u^1(\lambda) \le u^0(\mu) - u^0(\lambda).
\end{align}
If $u^1(\mu) > u^1(\lambda)$, then $u^0(\mu) > u^0(\lambda)$, which contradicts step 2. If $u^0(\mu) < u^0(\lambda)$, then $u^1(\mu) < u^1(\lambda)$, again contradicting step 2. Hence $u^1(\mu) \le u^1(\lambda)$ and $u^0(\mu) \ge u^0(\lambda)$, i.e., $\thelam[\mu]^1 \le \thelam^1$ and $\thelam[\mu]^0 \ge \thelam^0$. \qed

\subsection{Proof of \texorpdfstring{\cref{prop:threshold_effect_dm}}{Proposition \getrefnumber{prop:threshold_effect_dm}}}

In this proof we consider a fixed group $g$, and for readability, omit $g$ from $\alpha$, $p$, $\C_0$, $\C_1$, $\C_\star$ and $\Gamma_d$, and write $\theta := \theta^g$ and $\thelam[0] := \thelam[0]^g$. 

We can write the expected cost of a policy $\theta$ for a DS with score $x$ as:
\begin{align}
    c_\theta(x) := \C_1(x) \mathbf{1}\{x \ge \theta\} + \C_0(x) \mathbf{1}\{x < \theta\}
\end{align}
Under the selective dynamics (\cref{as:selective_dyn}), the post-deployment DM loss on group $g$ is:
\begin{align}
\label{eq:post_dm_group}
    \lpostg[\theta]{g} = \int_{\theta}^{\infty} \left[ \alpha(x) c_\theta(x + \bp) + (1 - \alpha(x)) c_\theta(x + \bm) \right] p(x) \, dx + \int_{-\infty}^{\theta} c_\theta(x + r) p(x) \, dx.
\end{align}
From the optimality of $\thelam[0]$, we can write $C_\star(x) = C_1(z) \mathbf{1}\{ x \ge \thelam[0] \} +C_0(z) \mathbf{1}\{ x < \thelam[0] \}$. Then, we can write the cost of a policy $\theta$ with respect to $\thelam[0]$. They disagree on decisions for scores in $J_\theta := [ \min \{\theta, \thelam[0] \}, \max \{ \theta, \thelam[0]\}]$, and:
\begin{align}
    c_\theta(x) = C_\star(x) + R_\theta(x) \qquad R_\theta(x) := \eta | \alpha(x) - \overline c | \mathbf{1}\{x \in J_\theta\}
\end{align}
where $R_\theta(x)$ is the regret term, the additional costs of not selecting the decision with minimal costs, which is $|C_0(x) - C_1(x)|$. 

Substituting $c_\theta(x)$ into \cref{eq:post_dm_group} and using the definition of $\Gamma_d$ (\cref{eq:gamma}), we have $\lpostg[\theta]{g} = M(\theta) + E(\theta)$ with
\begin{align}
    M(\theta) &:= \int_{-\infty}^{\theta} \Gamma_0(x) p(x) \, dx + \int_{\theta}^{\infty} \Gamma_1(x) p(x) \, dx, \\
    E(\theta) &:= \int_{\theta}^{\infty} \left[ \alpha(x) R_\theta(x + \bp) + (1 - \alpha(x)) R_\theta(x + \bm) \right] p(x) \, dx + \int_{-\infty}^{\theta} R_\theta(x + r) p(x) \, dx.
\end{align}
$M$ is the loss the DM would face if, after deployment, it took the optimal decision for every DS, and $E \ge 0$ is the regret from re-applying $\theta$ instead.

As $\alpha \in [0, 1]$, $\C_0$ and $\C_1$ are bounded and continuous, and so is $\C_\star = \min\{\C_0, \C_1\}$. Hence $\Gamma_0$ and $\Gamma_1$ are bounded and continuous, and since $p$ is integrable and continuous, the derivative of $M$ is:
\begin{align}
    M'(\thelam[0]) = \left( \Gamma_0(\thelam[0]) - \Gamma_1(\thelam[0]) \right) p(\thelam[0]) = \Delta_\Gamma(\thelam[0]) p(\thelam[0]).
\end{align}

To obtain the derivative of $E(\theta)$, we need to handle the term $\mathbf{1}\{x \in J_\theta\}$. Note that $R_{\thelam[0]}(x) = 0$ for all $x$, then $E(\thelam[0]) = 0$. Let $h := |\theta - \thelam[0]|$ and $\omega(h) := \sup_{|x - \thelam[0]| \le h} |\alpha(x) - \overline c|$; as $\alpha(\thelam[0]) = \overline c$ and $\alpha$ is continuous, $\omega(h) \to 0$ when $h \to 0$. Using $\alpha(x) \le 1$, $1 - \alpha(x) \le 1$, and that, for each shift $b$, the set $\{x : x + b \in J_\theta\}$ is an interval of length $h$:
\begin{align}
    0 \le E(\theta) \le \eta\, \omega(h) \sum_{b \in \{\bp, \bm, r\}} \int \mathbf{1}\{x + b \in J_\theta\} p(x) \, dx \le 3 \eta \lVert p \rVert_\infty \, h \, \omega(h),
\end{align}
where $\lVert p \rVert_\infty < \infty$ by \cref{as:outcome_model}. Therefore $|E(\theta) - E(\thelam[0])| / h \le 3 \eta \lVert p \rVert_\infty \omega(h) \to 0$, i.e., $E$ is differentiable at $\thelam[0]$ with $E'(\thelam[0]) = 0$. Combined with the derivative of $M$, $\lpostg[\theta]{g}$ is differentiable at $\thelam[0]$ with derivative $\Delta_\Gamma(\thelam[0]) p(\thelam[0])$.

We now prove the relation between the sign of the derivative and the change in $\lpost$. By \cref{as:outcome_model}, $p(\thelam[0]) > 0$, so the derivative has the sign of $\Delta_\Gamma(\thelam[0])$. If $\Delta_\Gamma(\thelam[0]) > 0$, by the definition of the derivative there is $\varepsilon > 0$ such that $(\lpostg[\theta]{g} - \lpostg[{\thelam[0]}]{g}) / (\theta - \thelam[0]) > 0$ for all $0 < |\theta - \thelam[0]| < \varepsilon$, i.e., $\lpostg[\theta]{g} < \lpostg[{\thelam[0]}]{g}$ for $\theta \in (\thelam[0] - \varepsilon, \thelam[0])$ and $\lpostg[\theta]{g} > \lpostg[{\thelam[0]}]{g}$ for $\theta \in (\thelam[0], \thelam[0] + \varepsilon)$. The case $\Delta_\Gamma(\thelam[0]) < 0$ is symmetric.

Lastly, we obtain the expression of $\Delta_\gamma(\thelam[0]^g)$ for each definition of DM costs. Since $\bm \le 0 \le \bp$, $r \le 0$ and $\alpha$ is increasing, $\alpha^-, \alpha^r \le \overline c \le \alpha^+$. Thus the optimal decision accepts at $\thelam[0] + \bp$ and rejects at $\thelam[0] + \bm$ and $\thelam[0] + r$, and, with $\alpha(\thelam[0]) = \overline c$,
\begin{align}
\label{eq:gap_general}
    \Delta_\Gamma(\thelam[0]) = \C_0(\thelam[0] + r) - \overline c\, \C_1(\thelam[0] + \bp) - (1 - \overline c)\, \C_0(\thelam[0] + \bm).
\end{align}
For $\Csel$, $\eta = c_{FP} - c_{TP}$, $\overline c = c_{FP} / \eta$, $\C_0 = 0$ and $\C_1(z) = c_{FP} - \eta \alpha(z) = \eta(\overline c - \alpha(z))$. Then \cref{eq:gap_general} gives
\begin{align}
    \Delta_\Gamma(\thelam[0]) = \overline c \eta (\alpha^+ - \overline c) = c_{FP}(\alpha^+ - \overline c).
\end{align}
For $\Cacc$, $\eta = c_{FN} + c_{FP}$, $\overline c = c_{FP}/\eta$, $\C_0(z) = c_{FN}\alpha(z)$ and $\C_1(z) = c_{FP}(1 - \alpha(z))$. Then \cref{eq:gap_general} gives
\begin{align}
    \Delta_\Gamma(\thelam[0]) &= c_{FN} \alpha^r - \overline c c_{FP} (1 - \alpha^+) - (1 - \overline c) c_{FN} \alpha^- \\
    &= c_{FP}\overline c(\alpha^+ - \overline c) + c_{FN}(1 - \overline c)(\overline c - \alpha^-) - c_{FN}(\overline c - \alpha^r) + \overline c \left[ c_{FN} \overline c - c_{FP}(1 - \overline c) \right],
\end{align}
where the last term is zero because $\overline c = c_{FP}/\eta$ and $1 - \overline c = c_{FN}/\eta$.

\subsection{Proof of \texorpdfstring{\cref{prop:post_fairness}}{Proposition \getrefnumber{prop:post_fairness}}}

As in the previous proof, we fix a group $g$ and omit it from $\alpha$, $p$, $\beta$, $\kappa_d$ and $\Delta_\kappa$.

We first prove when $\Delta_\kappa (x) \ge 0$. With rejection costs $\Brej$, we have that $\Delta_\kappa (x) = \dsc \alpha(x) \ge 0$ as $\dsc > 0$. With false negative costs $\Bfn$, we have that $\Delta_\kappa (x) = \dsc \left ( \alpha(x + r) - (1 - \alpha(x))(\alpha(x + \bp )\right )$, where $r \ge \bm$ is a sufficient condition to satisfy that $\Delta_\kappa(x) \ge 0$. 

For the first item, under \cref{as:selective_dyn}, a rejected DS remains rejected with expected cost $\kappa_0(x) = \beta(x + r)$, an accepted DS who repays remains accepted with no cost, and an accepted DS who defaults, with probability $1 - \alpha(x)$, has expected cost $\beta(x + \bm)$ if $x + \bm < \theta$ and no cost otherwise. Therefore, the post-deployment DS loss is:
\begin{align}
    \upost{g} = \int_{-\infty}^{\theta} \kappa_0(x) p(x) \, dx + \int_{\theta}^{\theta - \bm} \kappa_1(x) p(x) \, dx,
\end{align}
and we have:
\begin{align}
    \upost{g} - \upost[\phi]{g} &= \int_{\phi}^{\theta} \kappa_0(x) p(x) \, dx + \int_{\theta}^{\theta - \bm} \kappa_1(x) p(x) \, dx - \int_{\phi}^{\phi  -\bm} \kappa_1(x) p(x) \, dx \\
    &= \int_{\phi}^{\theta} \Delta_\kappa(x) p(x) \, dx + \int_{\phi  - \bm }^{\theta - \bm} \kappa_1(x) p(x) \, dx,
\end{align}
where the second equality follows by writing $K(t) := \int_{-\infty}^{t} \kappa_1(x) p(x) \, dx$, so that
\begin{align}
\int_{\theta}^{\theta - \bm} \kappa_1 p(x) \, dx - \int_{\phi}^{\phi -  \bm} \kappa_1 p(x) \, dx = \left( K(\theta - \bm) - K(\phi - \bm) \right) - \left( K(\theta) - K(\phi) \right)
\end{align}
The first term compares the DS in $[\phi, \theta)$, which are rejected by $\theta$ but accepted by $\phi$, and the second term reflects that the window of defaulters that fall below the threshold, $[\theta, \theta -\bm)$, shifts with the threshold. As $\kappa_1 \ge 0$, the second term is non-negative. The first term is non-negative whenever $\Delta_\kappa \ge 0$ on $[\phi, \theta]$, and then $\upost[\phi]{g} \le \upost{g}$.

For the second item, let $0 \le \lambda < \mu \le 1$. With Rawlsian fairness, \cref{prop:threshold} gives $\thelam[\mu]^g \le \thelam^g$ for both groups. Applying the first item with $\phi:= \thelam[\mu]$ and $\theta := \thelam$ (the case $\thelam[\mu]^g = \thelam^g$ is trivial), we have $\upost[{\thelam[\mu]}]{g} \le \upost[\thelam]{g}$ for both groups, and as the maximum is non-decreasing in each argument,
\begin{align}
    \frwpost[{\thelam[\mu]}] = \max_g \upost[{\thelam[\mu]}]{g} \le \max_g \upost[\thelam]{g} = \frwpost[\thelam].
\end{align}
With egalitarian fairness, \cref{prop:threshold} gives $\thelam[\mu]^1 \le \thelam^1$ and $\thelam[\mu]^0 \ge \thelam^0$. Applying the first result to group $1$ with $\phi := \thelam[\mu]$, $\theta := \thelam$ and to group $0$ with $\phi := \thelam$, $\theta := \thelam[\mu]$, we have $\upost[{\thelam[\mu]}]{1} \le \upost[\thelam]{1}$ and $\upost[\thelam]{0} \le \upost[{\thelam[\mu]}]{0}$, i.e., the fairer policy lowers the post-deployment loss of the disadvantaged group and raises the one of the advantaged group. Then, using $\upost[{\thelam[\mu]}]{1} \ge \upost[{\thelam[\mu]}]{0}$,
\begin{align}
    \fegpost[{\thelam[\mu]}] = \upost[{\thelam[\mu]}]{1} - \upost[{\thelam[\mu]}]{0} &\le \upost[\thelam]{1} - \upost[\thelam]{0} \\ &\le \left| \upost[\thelam]{1} - \upost[\thelam]{0} \right| = \fegpost[\thelam].
\end{align}
\qed

\subsection{Formal Statement of \texorpdfstring{\cref{theo:main}}{Theorem \getrefnumber{theo:main}} and Proof}
\label{app:final_theo}

\begin{theorem}
\label{thm:dominance}
Let the DM cost be $\Csel$ or $\Cacc$, the DS cost be $\Brej$ or $\Bfn$, and the fairness be $\Frw$ or $\Feg$, and write $V(\lambda) := \lpost[\thelam]$ for the post-deployment DM loss of the weight $\lambda$. Let $\overline\lambda \in (0, 1]$ be such that the conditions of the second part of \cref{prop:post_fairness} hold for every pair $0 \le \lambda < \mu \le \overline\lambda$, i.e., $\Delta_\kappa(\cdot, g) \ge 0$ on the interval between $\thelam[\mu]^g$ and $\thelam^g$ for both groups (which always holds for $\Brej$ and, for $\Bfn$, if $r \ge \bm$) and, for $\Feg$, $\upost[{\thelam[\mu]}]{1} \ge \upost[{\thelam[\mu]}]{0}$ (the groups do not swap places after deployment).
\begin{enumerate}
    \item If $V(\mu_0) < V(0)$ for some $\mu_0 \in (0, \overline\lambda]$, then there exists $\lambda^\dagger \in (0, \mu_0]$ such that every $\lambda \in [0, \lambda^\dagger)$ is post-deployment dominated by $\mu_0$.
    \item With Rawlsian fairness, if the disadvantaged group is strictly so, $\upre[{\thelam[0]}]{1} > \upre[{\thelam[0]}]{0}$, and $\Delta_\Gamma(\thelam[0]^1, 1) > 0$, then $V(\lambda) < V(0)$ for every sufficiently small $\lambda > 0$, so the condition of the first part holds.
    \item With egalitarian fairness, the condition of the first part reads
    \begin{align}
        \underbrace{\rho^1 \left( \lpostg[{\thelam[\mu_0]^1}]{1} - \lpostg[{\thelam[0]^1}]{1} \right)}_{\text{change on the disadvantaged group}} + \underbrace{\rho^0 \left( \lpostg[{\thelam[\mu_0]^0}]{0} - \lpostg[{\thelam[0]^0}]{0} \right)}_{\text{change on the advantaged group}} < 0.
    \end{align}
\end{enumerate}
\end{theorem}

\begin{proof}
We keep the notation of the proof of \cref{prop:threshold}: $u^g(\lambda) := \upre[\thelam]{g}$, the group-loss coordinates $v = (v^0, v^1)$, $\ell^g$ and $\Lo(v)$. As each DS pays a cost of at most $\dsc_{\max} := \max_{d, y} \B_{dy}$, $\upre{g} \in [0, \dsc_{\max}]$, and so $\Frw, \Feg \in [0, \dsc_{\max}]$. For both DM costs, $\overline c > 0$, so \cref{lemma:loss_coordinates} applies with $\Bfn$. Since $\rho^g$ does not depend on $\theta$ (\cref{as:selective_dyn}), the post-deployment DM loss decomposes as $\lpost = \sum_g \rho^g \lpostg[\theta^g]{g}$.

We first show that $\thelam^g \to \thelam[0]^g$ for both groups as $\lambda \to 0$. For $\lambda \in (0, 1)$, comparing the scalarized objective of $\thelam$ and $\thelam[0]$ gives $(1 - \lambda)(\lpre[\thelam] - \lpre[{\thelam[0]}]) \le \lambda (\fpre[{\thelam[0]}] - \fpre[\thelam]) \le \lambda \dsc_{\max}$. As $\thelam[0]^g$ minimizes $\lpreg[\theta^g]{g}$, each term of $\lpre[\thelam] - \lpre[{\thelam[0]}] = \sum_g \rho^g (\lpreg[\thelam^g]{g} - \lpreg[{\thelam[0]^g}]{g})$ is non-negative, so
\begin{align}
    0 \le \lpreg[\thelam^g]{g} - \lpreg[{\thelam[0]^g}]{g} \le \frac{\lambda \dsc_{\max}}{(1 - \lambda) \rho^g} \xrightarrow[\lambda \to 0]{} 0.
\end{align}
From the proof of \cref{lemma:loss_coordinates}, $\lpreg[\theta^g]{g}$ is continuous, strictly decreasing for $\theta^g \le \thelam[0]^g$ and strictly increasing for $\theta^g \ge \thelam[0]^g$. Then, for every $\delta > 0$, there is $m_\delta > 0$ such that $|\theta^g - \thelam[0]^g| \ge \delta$ implies $\lpreg[\theta^g]{g} - \lpreg[{\thelam[0]^g}]{g} \ge m_\delta$, and so $\thelam^g \to \thelam[0]^g$. By \cref{prop:threshold_effect_dm}, $\lpostg[\theta^g]{g}$ is differentiable, hence continuous, at $\thelam[0]^g$, and therefore $V(\lambda) \to V(0)$ as $\lambda \to 0$.

For the first part, as $V(\lambda) \to V(0) > V(\mu_0)$, there exists $\lambda^\dagger \in (0, \mu_0]$ such that $V(\lambda) > V(\mu_0)$ for all $\lambda \in [0, \lambda^\dagger)$. For such $\lambda$, we have $\lambda < \mu_0 \le \overline\lambda$, so \cref{prop:post_fairness} gives $\fpost[{\thelam[\mu_0]}] \le \fpost[\thelam]$. Together with $V(\mu_0) < V(\lambda)$, $\mu_0$ post-deployment dominates $\lambda$ (\cref{def:dominance}).

For the second part, we show that, for small $\lambda > 0$, only the threshold of the disadvantaged group moves, and that it moves strictly below $\thelam[0]^1$. First, as $\thelam^1 \to \thelam[0]^1$ and $\upre[\theta]{1}$ is continuous, $u^1(\lambda) \to u^1(0) > u^0(0)$, so $u^1(\lambda) > u^0(0)$ for small $\lambda$. Suppose $u^0(\lambda) \neq u^0(0)$; by \cref{prop:threshold}, $u^0(\lambda) < u^0(0)$. Let $v := (u^0(0), u^1(\lambda))$. Then $\Frw(v) = u^1(\lambda) = \Frw(u(\lambda))$, while $\Lo(v) < \Lo(u(\lambda))$ because $\ell^0$ is strictly decreasing on $[u^0(\lambda), u^0(0)]$ (\cref{lemma:loss_coordinates}), which contradicts \cref{eq:opt_consequence}. Hence $\thelam^0 = \thelam[0]^0$ for small $\lambda$.

Next, let $\lambda \in (0, 1)$ and suppose $u^1(\lambda) = u^1(0)$ (by \cref{prop:threshold}, $u^1(\lambda) \le u^1(0)$). For $\varepsilon \in (0, u^1(0) - u^0(0))$, let $v_\varepsilon := (u^0(\lambda), u^1(0) - \varepsilon)$, which lies in $I^0 \times I^1$ because $I^1$ is an interval with $\inf I^1 = 0$. As $u^0(\lambda) \le u^0(0) < u^1(0) - \varepsilon$, we have $\Frw(v_\varepsilon) = \Frw(u(\lambda)) - \varepsilon$. From the derivative in \cref{lemma:loss_coordinates}, $(\ell^1)'(u^1(0)) = \eta(\alpha(\thelam[0]^1, 1) - \overline c) / \beta(\thelam[0]^1, 1) = 0$, as $\beta(\thelam[0]^1, 1) = \dsc_{TN}(1 - \overline c) + \dsc_{FN}\, \overline c$ is positive. Hence $\Lo(v_\varepsilon) - \Lo(u(\lambda)) = \rho^1 (\ell^1(u^1(0) - \varepsilon) - \ell^1(u^1(0))) = o(\varepsilon)$, and the scalarized objective changes by
\begin{align}
    (1 - \lambda) \left( \Lo(v_\varepsilon) - \Lo(u(\lambda)) \right) + \lambda \left( \Frw(v_\varepsilon) - \Frw(u(\lambda)) \right) = (1 - \lambda)\, o(\varepsilon) - \lambda \varepsilon < 0
\end{align}
for small $\varepsilon$, which contradicts the optimality of $\thelam$. Hence $\thelam^1 < \thelam[0]^1$ for every $\lambda \in (0, 1)$.

As $\Delta_\Gamma(\thelam[0]^1, 1) > 0$, \cref{prop:threshold_effect_dm} gives $\varepsilon > 0$ such that $\lpostg[\theta^1]{1} < \lpostg[{\thelam[0]^1}]{1}$ for $\theta^1 \in (\thelam[0]^1 - \varepsilon, \thelam[0]^1)$. Since $\thelam^1 < \thelam[0]^1$ and $\thelam^1 \to \thelam[0]^1$, $\thelam^1$ lies in this interval for small $\lambda > 0$, and, as $\thelam^0 = \thelam[0]^0$,
\begin{align}
    V(\lambda) - V(0) = \rho^1 \left( \lpostg[\thelam^1]{1} - \lpostg[{\thelam[0]^1}]{1} \right) < 0.
\end{align}
Any such $\lambda \le \overline\lambda$ can be taken as $\mu_0$ in the first part.

The third part is the decomposition $V(\mu_0) - V(0) = \sum_g \rho^g (\lpostg[{\thelam[\mu_0]^g}]{g} - \lpostg[{\thelam[0]^g}]{g})$, where, by \cref{prop:threshold}, the threshold of group $1$ decreases and the threshold of group $0$ increases.
\end{proof}

\subsection{Extension to Different Costs for Rejections}
\label{app:general_rejection}

In the main text, $\Brej$ has the same cost for all rejections. We now consider the DS cost matrix
\begin{align}
    \B = \begin{pmatrix} \dsc_{TN} & \dsc_{FN} \\ 0 & 0 \end{pmatrix}
\end{align}
with $\dsc_{TN}, \dsc_{FN} > 0$, which penalizes rejections with $y = 0$ and $y = 1$ differently and, for $\dsc_{TN} \neq \dsc_{FN}$, does not recover demographic parity when combined with $\Feg$. The expected cost of rejection is $\beta(x, g) = \dsc_{TN}(1 - \alpha(x, g)) + \dsc_{FN}\,\alpha(x, g)$, and $\kappa_d$ and $\Delta_\kappa$ are defined as in \cref{eq:kappa}. As the proofs above are written in terms of $\dsc_{TN}$, $\dsc_{FN}$ and $\Delta_\kappa$, \cref{prop:threshold}, the two items of \cref{prop:post_fairness} and \cref{thm:dominance} hold for this matrix without change. The difference is in the sign of the DS acceptance gap, which is no longer guaranteed to be non-negative and which we can write as:
\begin{align}
    \Delta_\kappa(x, g) = {} & \dsc_{FN} \left[ \alpha(x + r, g) - (1 - \alpha(x, g))\,\alpha(x + \bm, g) \right] \nonumber \\
    & + \dsc_{TN} \left[ (1 - \alpha(x + r, g)) - (1 - \alpha(x, g))\,(1 - \alpha(x + \bm, g)) \right].
\end{align}
As $r \le 0$ and $\alpha$ is increasing, $1 - \alpha(x + r, g) \ge 1 - \alpha(x, g) \ge (1 - \alpha(x, g))(1 - \alpha(x + \bm, g))$, so the term multiplying $\dsc_{TN}$ is non-negative. We have two cases:
\begin{itemize}
    \item If $\dsc_{TN} \ge \dsc_{FN}$, we can replace $\dsc_{TN}$ by $\dsc_{FN}$ and, as the two brackets sum to $\alpha(x, g)$, obtain $\Delta_\kappa(x, g) \ge \dsc_{FN}\,\alpha(x, g) \ge 0$. The condition then always holds, as with $\Brej$.
    \item If $\dsc_{TN} < \dsc_{FN}$, the term multiplying $\dsc_{FN}$ can be negative, and the condition holds for all $(x, g)$ if $r \ge \bm$, as with $\Bfn$. When it fails, raising the threshold can reduce the post-deployment DS loss because the newly rejected DS become less qualified, and less qualified DS are less costly to reject.
\end{itemize}

\section{Implementation Details}
\label{app:implementation}

We implement a simulated environment where we first sample $500,000$ DS $(x_i, g_i, y_i)$ from the pre-deployment distribution $\D\pre$. Then, given a policy defined by thresholds $(\theta^0, \theta^1)$, we simulate sampling from the post-deployment distribution $\D(\theta)$ by applying the selective dynamics (\cref{as:selective_dyn}) to this same pool of DS, so the group of each DS is kept and only its score changes.

\begin{figure}[t]
    \centering
    \includegraphics[width=0.8\linewidth]{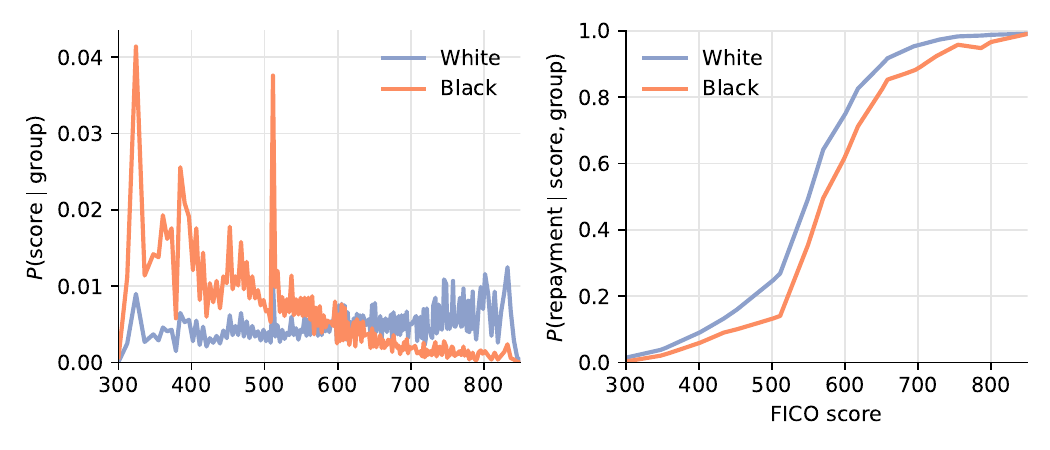}
    \caption{Empirical distribution of scores $p^g(x)$ and conditional probability of repayment $\alpha(x, g)$ estimated from FICO data to define the pre-deployment distribution $\D\pre$.}
    \label{fig:fico}
\end{figure}
\subsection{Simulation of Pre- and Post-Deployment Distributions}

Each DS has its group drawn as $g_i \sim \mathrm{Bern}(0.12)$ (12\% of Black applicants) and its score drawn from the empirical FICO score distribution of its group, $x_i \sim p^{g_i}(\cdot)$. The repayment probability $\alpha(x, g)$ is the empirical repayment rate for group $g$. FICO contains the repayment rates for $200$ scores, and we interpolate linearly to define this rate for the remaining scores in $[300, 850]$. Both $p^g(x)$ and $\alpha(x,g)$ are plot in \cref{fig:fico}.

The decision of DS $i$ is $\pi_\theta(x_i, g_i) = \mathbf{1}\{x_i \ge \theta^{g_i}\}$. A rejected applicant updates its score to $x_i + r$. An accepted applicant moves to $x_i + \bp$ if it repays and to $x_i + \bm$ if it defaults, which happens with probabilities $\alpha(x_i, g_i)$ and $1 - \alpha(x_i, g_i)$. Updated scores are clipped to the score range $[300, 850]$, so, e.g., an applicant who repays with $x_i + \bp > 850$ moves to $850$. To reduce sampling noise between different policies $\thelam$ and their induced distributions $\D(\thelam)$, we keep hidden for each DS the potential outcome. That is, in addition to having $(x_i, g_i, y_i)$ for each DS that represents the pre-deployment variables, we also keep $(x_i + r, g_i, y_i^r)$ that represents the updated score in the case of rejection, with the label sampled from $\alpha(x_i + r, g_i)$. For the acceptance case, we keep $(x_i + y_i \bp + (1 - y_i) \bm, g_i, y_i^a)$, where we update the score based on the repayment/default outcome ($y_i$), and we sample $y_i^a$ from this updated score. Then, when a policy accepts the applicant $i$, we reveal $(x_i + y_i \bp + (1 - y_i) \bm, g_i, y_i^a)$, and two policies that accept the same applicant will observe the same post-deployment label. The same is valid for rejection.

\subsection{Building the Pre-deployment PO Set}

To build the pre-deployment PO set $\mathcal S$, we select $\lambda$ linearly in the interval $[0, 1]$. Then, we optimize the scalarized objective (\cref{eq:scalarization}) by a grid search over all pairs of thresholds $(\theta^0, \theta^1)$. We evaluate $\theta^g$ for each unique value of $x$ in the pre-deployment distribution $\D\pre$, which, for FICO, has $200$ unique values. The DM loss and fairness objectives have different scales; therefore, weights in $[0, 1]$ may not cover much of the PO solution space. To handle this, we normalize the objectives by their range when evaluated at all thresholds. 

To quickly evaluate the objectives for each threshold $\theta^g$ and find the minimizer for each fairness weight $\lambda$, we precompute the DM loss and DS loss for each group across all evaluated thresholds and store them in an array. Then, evaluating the objective for a threshold $(\theta^0, \theta^1)$ only requires retrieving the value from the array. At $\lambda = 0$ and $\lambda = 1$, the minimizer of the scalarized objective may not be unique; in this case, we select among the minimizers of the DM loss (fairness) the one with the lowest fairness (DM loss), so that another minimizer does not dominate the selected policy. Finally, we deploy each minimizer $\thelam$ once on the pool of DS and evaluate both objectives with the same policy $\thelam$ on the induced post-deployment distribution $\D(\thelam)$. \cref{alg:experiment} summarizes the full procedure.

\begin{algorithm}[t]
\caption{Pre- and post-deployment evaluation of the scalarized set.}
\label{alg:experiment}
\begin{algorithmic}[1]
\Require Pool of DS $\{(x_i, g_i, y_i, y_i^r, y_i^a)\}_{i=1}^n$, set of weights $\Lambda \subset [0, 1]$, dynamics constants $\bp$, $\bm$, $r$
\State $T \gets \{x_i : i = 1, \dots, n\}$ \Comment{unique scores as candidate thresholds}
\For{each $t \in T$ and $g \in \{0, 1\}$}
    \State Compute $\lpreg[t]{g}$ and $\upre[t]{g}$ on the DS of group $g$
\EndFor
\For{each $\theta = (\theta^0, \theta^1) \in T \times T$}
    \State $L[\theta] \gets \sum_g \rho^g \lpreg[\theta^g]{g}$, \quad $F[\theta] \gets \F\left(\upre[\theta^0]{0}, \upre[\theta^1]{1}\right)$
\EndFor
\State $\tilde L \gets L / \left(\max_\theta L[\theta] - \min_\theta L[\theta]\right)$, \quad $\tilde F \gets F / \left(\max_\theta F[\theta] - \min_\theta F[\theta]\right)$ \Comment{normalize by range}
\For{each $\lambda \in \Lambda$}
    \State $\thelam \gets \argmin_{\theta \in T \times T} (1 - \lambda) \tilde L[\theta] + \lambda \tilde F[\theta]$
    \State Record pre-deployment objectives $\lpre[\thelam] = L[\thelam]$ and $\fpre[\thelam] = F[\thelam]$
    \For{$i = 1, \dots, n$} \Comment{sample from $\D(\thelam)$}
        \If{$x_i \ge \thelam^{g_i}$}
            \State $(x_i', y_i') \gets \left(x_i + y_i \bp + (1 - y_i) \bm, \; y_i^a\right)$
        \Else
            \State $(x_i', y_i') \gets \left(x_i + r, \; y_i^r\right)$
        \EndIf
    \EndFor
    \State Evaluate $\lpost[\thelam]$ and $\fpost[\thelam]$ with policy $\thelam$ on $\{(x_i', g_i, y_i')\}_{i=1}^n$
\EndFor
\State \Return $\left\{\left(\lpre[\thelam], \fpre[\thelam], \lpost[\thelam], \fpost[\thelam]\right)\right\}_{\lambda \in \Lambda}$
\end{algorithmic}
\end{algorithm}

\end{document}